\documentclass{article}

\usepackage{microtype}
\usepackage{stfloats}
\usepackage{subfigure}
\usepackage{graphicx}
\usepackage{subfigure}
\usepackage{booktabs} % for professional tables
\usepackage{pgfplots}
\usepackage{selectp}
\usepackage{hyperref}
\usepackage{amsmath, amssymb, amsfonts, mathrsfs, 
 amsthm, gensymb,hyperref, graphicx, url, color, accents}
\newtheorem{Theorem}{Theorem}
\newtheorem{Theorem*}{Theorem}

\newtheorem{Claim*}[Theorem]{Claim}

\newtheorem{CounterExample*}{$\overline{\hbox{\bf Example}}$}

\newtheorem{Example}[Theorem]{Example}
\newtheorem{Example*}[Theorem]{Example}

\newtheorem{Intuition*}[Theorem]{Intuition}
\newtheorem{Joke*}[Theorem]{Joke}
\newtheorem{Lemma}[Theorem]{Lemma}

\newtheorem{Lemma*}[Theorem]{Lemma}
\newtheorem{Open problem}[Theorem]{Open problem}

\newtheorem{Property}[Theorem]{Property}

\newtheorem{Question*}[Theorem]{Question}

\def \bSubexa    {\begin{subexa}}

\newcommand{\ignore}[1]{}
\newcommand{\fgnore}[1]{\vspace{#1}}

\newcommand{\EE}{\mathbb{E}}

\newcommand{\RR}{\mathbb{R}}

\newcommand{\reals}{\RR}

\def \cA     {{\cal A}}

\def \cC     {{\cal C}}
\def \cD     {{\cal D}}

\def \cI     {{\cal I}}

\def \cL     {{\cal L}}
\def \cM     {{\cal M}}
\def \cN     {{\cal N}}
\def \cO     {{\cal O}}
\def \cP     {{\cal P}}

\def \cR     {{\cal R}}

\def \cV     {{\cal V}}

\def \upto  {{,}\ldots{,}}

\def \Paren#1{{\left({#1}\right)}}

\newcommand{\ed}{\stackrel{\mathrm{def}}{=}}

\newcommand{\bi}{\begin{itemize}}
\newcommand{\ei}{\end{itemize}}
\def\orpro{\mathop{\mathchoice
   {\vee\kern-.49em\raise.7ex\hbox{$\cdot$}\kern.4em}
   {\vee\kern-.45em\raise.63ex\hbox{$\cdot$}\kern.2em}
   {\vee\kern-.4em\raise.3ex\hbox{$\cdot$}\kern.1em}
   {\vee\kern-.35em\raise2.2ex\hbox{$\cdot$}\kern.1em}}\limits}

\def\andpro{\mathop{\mathchoice
 {\wedge\kern-.46em\lower.69ex\hbox{$\cdot$}\kern.3em}
 {\wedge\kern-.46em\lower.58ex\hbox{$\cdot$}\kern.25em}
 {\wedge\kern-.38em\lower.5ex\hbox{$\cdot$}\kern.1em}
 {\wedge\kern-.3em\lower.5ex\hbox{$\cdot$}\kern.1em}}\limits}

\def\simge{\mathrel{%
   \rlap{\raise 0.511ex \hbox{$>$}}{\lower 0.511ex \hbox{$\sim$}}}}

\def\simle{\mathrel{
   \rlap{\raise 0.511ex \hbox{$<$}}{\lower 0.511ex \hbox{$\sim$}}}}

\newcommand{\sett}[1]{\{1\upto #1\}}

\newcommand{\norm}[1]{\lvert\lvert #1 \rvert\rvert}

\newcommand{\barI}{\bar{I}}

\newcommand{\barJ}{\bar{J}}

\newcommand{\barp}{\bar{p}}

\newcommand{\fspcs}[1]{f^{\textit{#1}}}
\newcommand{\tspcs}[1]{t^{\textit{#1}}}
\newcommand{\femp}{\fspcs{emp}}
\newcommand{\fest}{\fspcs{est}}
\newcommand{\fspl}{\fspcs{adj}}
\newcommand{\fout}{\fspcs{out}}
\newcommand{\fpoly}{\fspcs{poly}}

\newcommand{\fadls}{\fspcs{adls}}
\newcommand{\fsurf}{\fspcs{surf}}

\newcommand{\hatf}{\hat{f}}

\newcommand{\lone}[1]{\|#1\|_{1}}

\newcommand{\akk}[2]{\|#2\|_{\cA_{#1}}}
\newcommand{\akkint}[3]{\|#3\|_{\cA_{#1}, #2}}

\newcommand{\tv}[2]{d_{\tvv}(#1, #2)}
\newcommand{\loneint}[2]{{\|#2\|}_{1, #1}}

\newlength{\dhatheight}

\newcommand{\SURF}{\mathrm{SURF}}
\newcommand{\ADLS}{\mathrm{ADLS}}

\newcommand{\NADLS}{\mathrm{TURF}}
\newcommand{\TURF}{\mathrm{TURF}}

\newcommand{\VC}{\mathrm{VC}}

\newcommand{\barq}{\bar{q}}

\DeclareMathOperator{\Exp}{{\mathbb E}}
\newcommand{\dist}[2]{d\Paren{{#1},{#2}}}

\newcommand{\nlone}[2]{\|#1-#2\|_1}
\newcommand{\nloneint}[3]{\|#2-#3\|_{1,#1}}
\newcommand{\tvv}{\mathrm{TV}}

\newcommand{\parti}[3]{\overline{#1}^{#2,#3}}
\usepackage{xcolor}

\DeclareMathOperator*{\E}{\EE}

\usepackage[accepted]{icml2020}

\icmltitlerunning{TURF Algorithm}

\begin{document}

\twocolumn[
\icmltitle{TURF: A Two-factor, Universal, Robust, Fast\\
Distribution Learning Algorithm}




\icmlsetsymbol{equal}{*}

\begin{icmlauthorlist}
\icmlauthor{Yi Hao}{goo}
\icmlauthor{Ayush Jain}{goo}
\icmlauthor{Alon Orlitsky}{goo}
\icmlauthor{Vaishakh Ravindrakumar}{goo}
\end{icmlauthorlist}

\icmlaffiliation{goo}{Electrical and Computer Engineering, University of California, San Diego}

\icmlcorrespondingauthor{Vaishakh Ravindrakumar}{varavind@ucsd.edu}

\icmlkeywords{Information Theory, Computational Learning Theory, Statistics, Probability Estimation}

\vskip 0.3in
]



\printAffiliationsAndNotice{}  % leave blank if no need to mention equal contribution

\begin{abstract}
Approximating distributions from their samples is a canonical statistical-learning problem. One of its most powerful and successful modalities approximates every distribution to an $\ell_1$ distance essentially at most a constant times larger than its closest $t$-piece degree-$d$ polynomial, where $t\ge1$ and $d\ge0$. Letting $c_{t,d}$ denote the smallest such factor, clearly $c_{1,0}=1$, and it can be shown that $c_{t,d}\ge 2$ for all other $t$ and $d$. Yet current computationally efficient algorithms show only $c_{t,1}\le 2.25$ and the bound rises quickly to $c_{t,d}\le 3$ for $d\ge 9$. We derive a near-linear-time and essentially sample-optimal estimator that establishes $c_{t,d}=2$ for all $(t,d)\ne(1,0)$. Additionally, for many practical distributions, the lowest approximation distance is achieved by polynomials with vastly varying number of pieces. We provide a method that estimates this number near-optimally, hence helps approach the best possible approximation. Experiments combining the two techniques confirm improved performance over existing methodologies.\looseness=-1

% \ignore{
% Approximating distributions from their samples is a canonical statistical-learning problem.
% One of its most \tcb{significant} \tcr{powerful and successful} modalities approximates every distribution to an $\ell_1$ distance 
% at most $c$ times larger than its closest \tcb{piecewise} \tcr{$t$-piece degree-$d$} polynomial. 
% It is well known that for all algorithms $c\ge 2$, while all known general subexponential-time algorithms \tcr{guarantee} $c>3$. 
% We present a near-linear-time algorithm with $c$ that is arbitrarily close to the optimal 2.
% Additionally, for many practical distributions, the lowest approximation distance is achieved by polynomials with vastly varying number of pieces.
% We provide a method that estimates this number near-optimally, hence helps approach the best possible approximation. 
% Experiments combining the two techniques \tcr{confirm} improved performance over existing methodologies.%\looseness=-10
% }
% \newbla{In fact $\TURF$ is a construction that can  be applied to any 
% constant approximation factor proper estimator for the polynomials to 
% obtain the optimal $2$-factor, and is based on novel bounds 
% that relate the  $\ell_\infty$ and $\ell_1$ norm of 
% polynomials, which may be interesting in its own right.}

% \blu{ADLS: $3, O(n \log n), \sqrt{t(d+1)/n}$\\
% SURF: $2.25-3, O(n \log n), \sqrt{t(d+1)\log n/n}$ (no $\log n$ for $t=1$)\\
% TURF: $2, O(n \log n), \sqrt{t(d+1)/n}$\\
% Yat sel: $3, e^n, \sqrt{t(d+1)/n}$\\
% Bosquet: $2, e^n, \sqrt{t(d+1)\log n/n}$
% }
\end{abstract}

\section{Introduction}
\label{sec:introduction}
\fgnore{-.5em}
Learning distributions from samples is one of the oldest~\cite{pea95}, most natural~\cite{sil86}, and 
important statistical-learning paradigms~\cite{givens2012computational}.
Its numerous applications include
epidemiology~\cite{bithell1990application}, 
% traffic modeling~\cite{hashimoto2016development} \blu{consider replacing with GANs or something more significant}, 
economics~\cite{zambom2013review}, 
anomaly detection~\cite{pimentel2014review}, 
language based prediction~\cite{gerber2014predicting},
GANs~\cite{goodfellow2014generative},
and many more, as outlined in several books and surveys e.g.,~\cite{tukey1977exploratory, sco12, diakonikolas2016learning}.
% \newbla{Avoid usage of `oldest'. Instead check that Guassian best paper or Good Turing introduction for a good example.
% \url{https://proceedings.neurips.cc/paper/2018/file/70ece1e1e0931919438fcfc6bd5f199c-Paper.pdf} 
% Applications other than traffic modeling (not very relevant to ML). 
% ``Many learning applications 
%, ranging from language-processing staples such as speech recognition
%and machine translation to biological studies in virology and bioinformatics, 
% call for estimating large
% discrete distributions from their samples.''}

\fgnore{-.25em}

Consider estimating an unknown, real, discrete, continuous, or mixed distribution $f$
from $n$ independent samples $X^n:=X_1,...,X_n$ it generates.
%Let $f$ be an unknown continuous-discrete distribution over $\reals$.
%Consider estimating $f$ from $n$ independent samples $X^n:=$ $X_1,...,X_n$ it generates.
A \emph{distribution estimator} maps $X^n$ to an approximating distribution $\fest$
% \tcr{later $\fest(X^n)$?}\blu{good catch - this def was changed recently, will update} 
meant to approximate $f$.
We evaluate its performance via the expected \emph{$\ell_1$ distance} $\EE\nlone{\fest}{f}$.\looseness-10 %between $f$ and its estimate.

\fgnore{-.25em}

The $\ell_1$ distance between two functions $f_1$ and $f_2$, $\nlone{f_1}{f_2}:=\int_{\reals} |f_1-f_2|$, 
is one of density estimation's most common distance measures~\cite{dev12}.
Among its several desirable properties, its value remains unchanged under linear transformation of the underlying domain, 
and the absolute difference between the expected values of any bounded function of the observations under 
$f_1$ and $f_2$ is at most a constant factor larger than $\nlone{f_1}{f_2}$, 
as for any bounded $g: \reals \rightarrow \reals$, $\big|\E_{f_1}[g(X)]-\E_{f_2}[g(X)]\big|
\le \max_{x\in \reals} g(x) \cdot  \nlone{f_1}{f_2}$.
Further, a small $\ell_1$ distance between two distributions implies a small difference between
any given Lipschitz functions of the two distributions.
Therefore, learning in $\ell_1$ distance implies a bound on the
error of the plug-in estimator for 
Lipschitz functions of the underlying distribution
~\cite{hao2020unified}.

\fgnore{-.25em}

Ideally, we would like to learn any distribution to a small $\ell_1$ distance.
However, arbitrary distributions cannot be learned in $\ell_1$ distance with any number of samples~\cite{devroye1990no}, as the following example shows.

\fgnore{-.25em}

\begin{Example}
\label{example:1}
Let $u$ be the continuous uniform distribution over $[0,1]$.
For any number $n$ of samples, construct a discrete distribution $p$ by assigning probability $1/n^3$ to each of $n^3$ random points in $[0,1]$.
By the birthday paradox, $n$ samples from $p$ will be all distinct with high probability and follow the same uniform distribution as $n$ samples from $u$, and hence $u$ and $p$ will be indistinguishable. 
As $\lone{u-p}=2$, the triangle inequality implies that for any
estimator $ \fest $, 
$ \max_{f\in \{u, p\}}\EE\nlone{\fest}{f} \gtrsim 1$.
\end{Example}

\fgnore{-.25em}

A common remedy to this shortcoming assumes that the distribution $f$ belongs to a structured approximation class $\cC$, 
% such as log-concave, uni-modal mixtures, or piecewise polynomials. 
for example unimodal~\cite{birge87}, log-concave~\cite{dev12} and
Gaussian~\cite{acharya2014near,ash18} distributions.

\fgnore{-.25em}

The \emph{min-max learning rate} of $ \cC$ is the lowest worst-case expected distance achieved by any estimator,
\fgnore{-.25em}
\[
\cR_n(\cC) \ed
\min_{\fest} \max_{f\in \cC}\E_{X^{n}\sim f} \nlone{\fest_{X^n}}{f}.
\fgnore{-.25em}
\]
The study of $\cR_n(\cC)$ for various classes such as
Gaussians, exponentials,
and discrete distributions has been the focus of many works e.g.,~\cite{vapnik1999overview,kamath2015learning, han2015minimax, cohen2020learning}.

Considering all pairs of distributions in $\cC$, 
\cite{yatracos1985rates} defined a collection of subsets with VC dimension~\cite{vapnik1999overview} $\VC(\cC)$,
and applying the minimum distance estimation method~\cite{wolfowitz1957minimum}, showed that
\fgnore{-.25em}
\[
\cR_n(\cC) = \cO(\sqrt{\VC(\cC)/n}).
\fgnore{-.25em}
\]
However, real underlying distributions $f$ are unlikely to fall \emph{exactly} in any predetermined class.
Hence~\cite{yatracos1985rates} also considered approximating $f$ nearly as well as its best approximation in $\cC$.
Letting
\fgnore{-.5em}
\[
\nlone{f}{\cC}
\ed
\inf_{g\in \cC}\nlone{f}{g}
\fgnore{-.25em}
\]
be lowest $\ell_1$ distance between $f$ and any distribution in $\cC$, he designed an estimator $\fspcs{Yat}$, possibly outside $\cC$, whose $\ell_1$  distance from $f$ is close to $\nlone{f}{\cC}$.
For all distributions $f$,\looseness-1
\fgnore{-.25em}
\[
\E \nlone{\fspcs{Yat}}{f}
\le
3\cdot\nlone{f}{\cC}+\cO(\sqrt{\VC(\cC)/n}).
\fgnore{-.25em}
\]
For many natural classes, $\cR_n(\cC)=\Theta(\sqrt{\VC(\cC)/n})$. Hence, an estimator $\fest$ is called a $ c $\emph{-factor approximation} for $\cC$ if for any distribution $f$,
\fgnore{-.25em}
\[
\E\nlone{\fest}{f}\le c\cdot \nlone{f}{\cC} + \cO(\cR_n(\cC)).
\fgnore{-.25em}
\]
$\nlone{f}{\cC}$ and $\cO(\cR_n(\cC))$ may be thought of as the error's \emph{bias} and \emph{variance} components.

\fgnore{-.25em}

A small $c$ is desirable as it upper bounds the asymptotic error when $n\nearrow \infty$ for $f\notin \cC$,
hence providing robustness guarantees when the underlying distribution does not quite follow the assumed model. 
It ensures robust estimation also under the Huber contamination model~\cite{huber1992robust} where with probability $0\le \mu\le 1$, $f$ is perturbed by an arbitrary noise, and the error incurred by a $c$-factor approximation is upper bounded as $c\cdot \mu$. 

\fgnore{-.25em}

One of the more important distribution classes is the collection $\cP_{t,d}$ of $ t $-piecewise 
degree-$d $ polynomials.
For simplicity, we assume that all polynomials in $\cP_{t,d}$ are defined over a known interval $I\subseteq\reals$, hence 
any $p\in \cP_{t,d}$ consists of degree-$d$ polynomials $p_1\upto p_t$, each defined over one part in a partition $I_1\upto I_t$ of $I$.

The significance of $\cP_{t,d}$ stems partly from the fact that it approximates numerous important distributions even with small $t$ and $d$.
For example, for every distribution $f$ in the class $\cL$ of log-concave distributions,  $\nlone{f}{\cP_{t,1}} = \cO(t^{-2})$~\cite{chan14}.
Also, $\VC(\cP_{t,d})=t(d+1)$, e.g.,~\cite{jay17}.

It follows that if $\fest$ is a $c$-factor estimator for $\cP_{t,1}$, 
% \tcr{d?} 
then for all
$f\in \cL$, $\E \nlone{\fest}{f}
\le
c\cdot\nlone{f}{\cP_{t,1}}+\cO(\sqrt{t/n})=\cO(t^{-2})+\cO(\sqrt{t/n}).
$
Choosing $t=n^{1/5}$ to equate the bias and variance terms, 
$\fest$
% this constant-factor
achieves an expected $\ell_1$ error $\cO(n^{-2/5})$,
which is the optimal min-max learning rate of $\cL$~\cite{chan14}. \looseness-1

Lemma~\ref{lem:minor} in Appendix~\ref{pf:minor} shows a stronger result.
% If $\fest$ is a $c$-factor approximation for $\cP_{t,d}$ for some $t$ and $d$ and if the min-max rate of a distribution class $\cC$ is at-most the $c$-factor error of $\cP_{t,d}$, then $\fest$ is also a $c$-factor approximation for $\cC$.
If $\fest$ is a $c$-factor approximation for $\cP_{t,d}$ for some $t$ and $d$ and achieves the min-max rate of a distribution class $\cC$, then $\fest$ is also a $c$-factor approximation for $\cC$.
In addition to the log-concave class, this result also holds for Gaussian,
and unimodal distributions, 
and for their mixtures.
\fgnore{-.25em}
\section{Contributions}
\fgnore{-.5em}
\label{sec:contrib}
\textbf{Lower Bounds:}
As noted above, it is beneficial to find the smallest approximation factor for $\cP_{t,d}$.
The following simple example shows that if we allow sub-distributions, even simple collections may have an approximation factor of at least 2.
\begin{Example}
Let class $\cC$ consist of the uniform distribution $u(x)=1$ and the subdistribution $z(x)=0$, over $[0,1]$.
Consider any estimator $\fest$. 
Let $f_u=\fest$ when $X^n\sim u $ as $n\nearrow \infty$.
Since $\nlone{u}{\cC}=0$, for $\fest$ to achieve finite approximation factor, we must have  $\nlone{f_u}{u}=0$.
Now consider the discrete distribution $p$ in Example~\ref{example:1}.
Since its samples are indistinguishable from those of $u$, $\fest_{X^n}=f_u$ also for $X^n\sim p$.
But then $\nlone{f_u}{p}\ge \nlone{u}{p}-\nlone{f_u}{u}=2=2\cdot\nlone{p}{\cC}$, so $\fest$ has approximation factor $\ge 2$. \looseness-1
\end{Example}
Our definition however considers only strict distributions, complicating lower bound proofs.
Let $c_{t,d}$ be the lowest approximation factor for $\cP_{t,d}$.
$\cP_{1,0}$ consists of a single distribution over a known interval, hence $c_{1,0}=1$.
\cite{chan14} showed that for all $t\ge 2$ and $d\ge 0$, $c_{t,d}\ge 2$. 
% We extend this bound to  lower bound via the following lemma proved 
The following lemma, proved in Appendix~\ref{pf:example2}, shows that $c_{1,d}\ge 2$ for all $d\ge1$, 
and as we shall see later, establishes a precise lower bound for all $t$ and $d$.
\begin{Lemma}
\label{lem:example2}
For all $(t,d)$ except $(1,0)$, $c_{t,d}\ge 2$.
% Suppose $t\ge 2$ or $d\ge 1$. Then for any $\epsilon>0$, $n\ge 1$, 
% for all estimators
% $\fest_{X^n}$ there exists an $f$ such that
% \[
% \E\nlone{\fest_{X^n}}{f}
% \ge  
% (2-\epsilon)\cdot\nlone{f}{\cP_{t,d}}.
% \]
\end{Lemma}
\fgnore{-.25em}
\textbf{Upper Bounds:} 
As discussed earlier, $\fspcs{Yat}$ is a $3$-factor approximation for $\cP_{t,d}$.
However its runtime is $n^{\cO(t(d+1))}$.
For many applications, $t$ or $d$ may be large, and even increase with $n$, 
for example in learning unimodal distributions, we select 
$t=\cO(n^{1/3})$~\cite{birge87}, resulting in exponential time complexity.
\cite{chan14} improved the runtime to polynomial in $n$ independent of $t,d$, 
and~\cite{jay17} further reduced it to near-linear $ \cO(n\log n) $. 
\cite{hao2021surf} derived a $\cO(n\log n)$ time algorithm, SURF, achieving $c_{t,1}=2.25$, and $c_{t,d}<3$ for $d\le 8$.
They also showed that this estimator can be parallelized to run in time $ \cO(n\log n /t)$.
\cite{bousquet19,bousquet2021statistically}'s estimator for the improper learning setting (wherein $\fest$ can be any distribution as we consider in this paper)
achieves a bias nearly within a factor of $2$, but the variance term exceeds $\cO(\cR_n(\cP_{t,d}))$, hence does not satisfy the constant factor approximation definition.
Moreover, like Yatracos, they suffer a prohibitive $n^{\cO(t(d+1))}$ runtime, that could be exponential for some applications.

\fgnore{-.25em}

Our main contribution is an estimator, $\TURF$, \emph{a two factor, universal, robust and fast estimator} that achieves an approximation factor that is arbitrarily close to the optimal $c_{t,d}=2$ in near-linear $\cO(n\log n)$ time.
$ \TURF $
is also simple to implement as a step on top of the existing merge 
routine in~\cite{jay17}.
The construction of our estimate 
relies on upper bounding
the maximum absolute value of polynomials (see Lemma~\ref{lem:poly}) based on their $\ell_1$ norm, similar to the
Bernstein~\cite{rahman2002analytic} and Markov Brothers'~\cite{achieser1992theory}
inequalities. 
% We use these inequalities in Lemma~\ref{lem:poly} to upper bound the difference between the maximum and minimum values of polynomial in terms of its $ \ell_{1} $ norm. 
We show 
% for an absolute constant $c>0$,
for any $p\in \cP_{1,d}$ and $a\in[0,1)$,
\fgnore{-.25em}
\[
\|p\|_{\infty, [-a,a]}
\,\le\,
\frac{28(d+1)\loneint{[-1,1]}{p}}{\sqrt{1-a^2}},
\fgnore{-.25em}
\]
where $\|\cdot\|_{\cdot , I}$ indicates the respective norms over any interval $I\subseteq\reals$.
This point-wise inequality reveals a novel connection between the $\ell_\infty$ and $\ell_1$ norms of a polynomial, 
which may be interesting in its own right.

\fgnore{-.25em}

\textbf{Practical Estimation:} 
% For any collection of $c_{t,d}$-factor approximations for $\cP_d$, $\fest_{t,d}$, it is not 
For many practical distributions, the optimal parameters values of $t,d$ in approximating with $\cP_{t,d}$ many be unknown.
% for an arbitrary $f$. 
While for common structured classes such as Gaussian, log-concave and unimodal, and their mixtures,
it suffices to choose $d$ to be any small value, but the optimal choice of $t$ 
can vary significantly. For example, for any constant $d$, for a unimodal $f$, 
the optimal $t=\cO(n^{1/3})$ pieces whereas for a smoother log-concave $f$, 
significantly lower errors are obtained with a much smaller $t=\cO(n^{1/5})$. 
% Let $\fest_{t,d}$ be the estimate $\fest_{X^n}$ 
Given a family of $c_{t,d}$-factor approximate estimators for $ \cP_{t,d} $, $\fest_{t,d}$,
a suitable objective is to select the number of pieces, $\tspcs{est}_d$ to achieve for any given degree-$d$, 
\fgnore{-.25em}
\begin{align}
% \fgnore{-1em}
\label{eqn:cv}
\E \!\|\fest_{\tspcs{est}_{d}}\!\!-\!f\|_1 \! \!\le  \!\min_{t\ge 1}\!
\Paren{\!c_{t,d}\|{f}\!-\!{\cP_{t,d}}\|_1 \!\!+ \! \cO\Paren{\!\sqrt{t(d\!+\!1)/n}}\!}.
\fgnore{-.25em}
\end{align}
Simple modifications to existing  
cross-validation approaches~\cite{yatracos1985rates} partly achieve 
Equation~\eqref{eqn:cv} with the larger $c = 3 c_{t,d}$ along with an additive $\cO(\log n/\sqrt{n})$. 
Via a novel cross-validation 
technique, we obtain a $\tspcs{est}_d$
that satisfies Equation~\eqref{eqn:cv} with the factor $c$ arbitrarily close to the optimal $c_{t,d}$ with an additive
$\cO(\sqrt{\log n/n})$. In fact, this technique 
removes the need to know 
parameters beforehand in other related settings as well, such as
the \emph{corruption level} in robust estimation that all existing works assume is known.
We elaborate this in~\cite{JORcomm}.
% \newbla{upload Arxiv stuff at the end of this week to cite by the deadline}.

\fgnore{-.25em}

Our experiments reflect the improved errors of $ \TURF $ over existing algorithms 
in regimes where the bias dominates.

\fgnore{-.5em}
\section{Setup}
\fgnore{-.25em}
\label{sec:setup}
\subsection{Notation and Definitions}
\fgnore{-.25em}
Henceforth, for brevity, we skip the $X^n$ subscript when referring to estimators.
Given samples $X^{n}\sim f$, the \emph{empirical distribution} is defined via the dirac delta function $\delta(x)$ as
\[
\fgnore{-.5em}
\femp(x)
\ed \sum_{i=1}^{n}
\frac{\delta(x-X_i)}{n},
\]
\fgnore{-.25em}
allotting a $1/n$ mass at each sample location.

\fgnore{-.25em}

Note that if an estimator $g$ is partly negative 
but integrates to $1$,
then $g' \ed \max\{g,0\}/\int_\reals \max\{g,0\}$, satisfies $\tv{g'}{f}\le \tv{g}{f}$ for any distribution $f$, e.g.,~\citet{dev12}. 
This allows us to estimate
using any real normalized function as our estimator.

\fgnore{-.25em}

For any interval $ I\subseteq \reals $ and integrable functions
$ g_1, g_2:\reals\rightarrow \reals $, 
let 
$\nloneint{I}{g_1}{g_2}$
denote the $ \ell_1 $ distance
evaluated over $ I $.
Similarly, for any class $\cC$ of real functions, let $ \nloneint{I}{g}{\cC} $
denote the least $ \ell_1 $ distance between $g$ and members of $\cC$ over $I$. 

\fgnore{-.25em}

The $\ell_1$ distance between $f$ and $\fest$
is closely related to their $\tvv$ or statistical distance as 
\[
1/2\cdot \nlone{\fest}{f}
=\tv{\fest}{f} \ed\sup_{S\in \reals}\Big|\int_S \fest-f\Big |,
\]
the greatest absolute difference in areas of $\fest$ and $f$ over all subsets of $\reals$.
As we argued in the introduction, a direct approach to estimate 
$f$ in all possible subsets of $\reals$ is not feasible with finitely many samples.
Instead, for a given $k\ge 1$, the $\cA_k$ distance~\cite{dev12}
considers the largest difference between $f$ and 
$\fest$ on real subsets with at most $k$ intervals.
As we show in Lemma~\ref{lem:vc}, it is 
possible to learn any $f$ in $\cA_k$ distance simply by 
using the empirical distribution $\femp$.

We formally define the $\cA_k$ distance as follows.
For any given $k\ge 1$ and interval $ I\subseteq \reals $, 
let $\cI_k(I)$ be the set of all unions 
of at most $k$ intervals contained in $I$.
Define the $\cA_k$ distance between $g_1, g_2$ as
\fgnore{-.25em}
\[
\akkint{k}{I}{g_1-g_2}\ed \sup_{S\in\cI_k(I)}
|g_1(S)-g_2(S)|,
\fgnore{-.25em}
\]
where $g(S)$ denotes the area of 
the function $g$ on the set $S$.
For example, if $ I=[0,1] $
and $ g_1(x) = x, \ g_2(x)=2/3$, the $\cA_1$ distance $\akkint{1}{I}{g_1-g_2} = \int_{0}^{2/3}|z-2/3|dz= 2/9$.
Suppose $I$ is the support of $f$. Use this to define $ \akk{k}{g_1-g_2} \ed \akkint{k}{I}{g_1-g_2} $.

\fgnore{-.25em}

For two \emph{distributions} $g_1$ and $g_2$,
the $\cA_k$ distance 
% between two distributions 
is at-most half the $\ell_1$ distance and with equality achieved as $k\nearrow \infty$ since $\cI_k(I)$ approximates all subsets of $I$ for large $k$,
\fgnore{-.25em}
\[
\akkint{k}{I}{g_1-g_2}\le
% \tvi{I}{g_1}{g_2} = 
1/2\cdot \nlone{g_1}{g_2}.
\fgnore{-.25em}
\] 
The reverse is not true, the $\cA_k$ distance between two functions may be made arbitrarily small even for a constant $\ell_1$ distance.
For example the $\ell_1$ distance between any distribution $f$ and its 
empirical distribution $\femp$ is $2$ for any $n\ge 2$. However, the $\cA_k$ distance between $f$ 
and $\femp$ goes to zero. 
% for any distribution $f$. 
The next lemma, which is a consequence of VC 
inequality~\cite{dev12}, gives the rate at which 
$\akk{k}{\femp - f }$ goes to zero. 
%For example it is known that over any finite 
%collection of intervals the empirical distribution $ \femp $ and $ f $ largely agree in their probability 
%masses. The following VC Inequality~\cite{dev12} upper bounds 
%the $ \cA_{k} $ distance between $ 
%\femp $ and $ f $ even as $\lone{\femp -f }=2$.
% \fgnore{-.25em}
\begin{Lemma}
\label{lem:vc}
\cite{dev12}
Given $X^{n}\sim f$ according to any real distribution $f$, 
% w.p. $\ge 1-\delta$,
% the $\cA_k$ distance between $\femp$ and $f$,
\[ 
\E \akk{k}{\femp - f }=  \cO\Paren{\sqrt{{k}/{n}}}.
\fgnore{-1em}
\]
\fgnore{-1em}
\end{Lemma}
Note that if $f$ is a discrete distribution with support size $k$, Lemma~\ref{lem:vc} implies $\Exp\nlone{\femp}{f}= \Exp\akk{k}{\femp - f }\le \cO(\sqrt{{k}/{n}})$, matching the rate of learning discrete distributions. 
Since arbitrary continuous distributions can be thought 
of as infinite dimensional discrete distributions where $k\rightarrow \infty$, the lemma does not bound this error.
% $ k $-disjoint-intervals over $ I $.
% The \emph{$\cA_k$ norm} of $g$ over $I$ is defined as
% Given an interval $I\subseteq \reals$, the $ \cA_{k}$ norm of $ g $ is the largest sum of the absolute value of the signed areas of $g$ over $k$ disjoint \emph{intervals} over $I$. Formally,
% \[
% \akkint{k}{I}{g}\ed \sup_{\Paren{I_1\upto I_k}:I_{i}\subseteq I, \ I_{i}\cap I_{j}=\phi,  \forall\, i,j\in[k]} \sum_{j\in[k]}\Big |\int_{I_j} g(z)dz\Big |.
% \]
% where we recall that $ g(J) $ the measure of $ g $ in $ J\subseteq \reals $.
% In other words, the $ \cA $ norm corresponds to the largest sum
% of absolute values of the measure of a function over any set 
% of $ k $ disjoint intervals. 
\fgnore{-.25em}
\subsection{Preliminaries}
\fgnore{-.25em}
The following $\cA_k$-distance properties are helpful. 
\begin{Property}
\label{prop:1}
\fgnore{-.25em}
Given the partition $I_1, I_2$ of any interval $I\subseteq \reals$ integrable functions $g_1,g_2\in I$,
and integers $ k_{1}, k_{2}$,
\fgnore{-.25em}
\[\akkint{k_1}{I_{1}}{g_1-g_2}+\akkint{k_2}{I_{2}}{g_1-g_2}
\le 
\akkint{k_{1}+k_{2}}{I}{g_1-g_2}.
\fgnore{-.25em}
\]
\end{Property}
Property~\ref{prop:1} follows since the interval 
choices with $k_1$ and $k_2$ intervals respectively 
that achieve the suprema of $\akkint{k_1}{I_1}{g_1-g_2} $ and $\akkint{k_2}{I_2}{g_1-g_2}$ are included in 
the $k_1+k_2$ interval partition considered in the RHS.
\begin{Property}
\fgnore{-.25em}
\label{prop:2}
Given any interval $I\subseteq \reals$, 
integrable functions $g_1,g_2\in I$, 
and integers $ k_{1}\ge k_{2}> 0 $,
\fgnore{-.25em}
\[ \akkint{k_1}{I}{g_1-g_2}\le \frac{k_{1}}{k_{2}}\cdot \akkint{k_{2}}{I}{g_1-g_2}. 
\fgnore{-.25em}\]
\end{Property}
Property~\ref{prop:2} follows from selecting the $k_2$ intervals 
with the largest contribution to $\akkint{k_1}{I}{g_1-g_2}$ in the RHS expression,
among the $k_1$ interval partition that attains $\akkint{k_1}{I}{g_1-g_2}$ on the LHS.
In Sections~\ref{sec:singlep},~\ref{sec:multip} that follow, we consider deriving the optimal rates of learning with piecewise polynomials.
\fgnore{-.5em}
\section{A 2-Factor Estimator for $\cP_{1,d}$}
\label{sec:singlep}
\fgnore{-.25em}
Our objective is to obtain a  $2$-factor approximation 
for the piecewise class, $\cP_{t,d}$. 
To achieve this, we first consider the single-piece class $\cP_{1,d}$ that for simplicity we denote by $\cP_d = \cP_{1,d}$,
and then use the resulting estimator as a sub-routine for the multi-piece class. 
\fgnore{-.5em}
\subsection{Intuition and Results}
\label{sec:above}
\fgnore{-.25em}
It is easy to show from the triangle inequality
that if an estimator is as close to \emph{all} degree-$d$ polynomials as their $\ell_1$ distances to $f$, then
the estimator achieves an $\ell_1$ distance to $f$ that is nearly twice that of the best degree-$d$ polynomial.
\fgnore{-.25em}

Let $|I|$ denote the length of an interval $I$. 
The \emph{histogram} of an integrable function $g$ over $I$  
is $\bar g_I\ed|\int_I g|/|I|$, where we assign 
zero to division by zero.

\fgnore{-.25em}

Let $\fpoly\in \cP_d$ be a polynomial estimator of $f$ over 
% an interval 
$I$, and let $\fspl$ be the function obtained by adding to $\fpoly$ a constant to match its mass to $f$ over $I$.
For any $p\in \cP_d$,
\fgnore{-.5em}
\begin{align*}
\nloneint{I}{\fspl}{p} &\overset{(a)}\le  \nloneint{I}{\fspl\!-\!p}{\Paren{\overline{\fspl}\!-\!\bar{p}}}\!+\! \|\overline{\fspl}\!-\!\bar{p}\|_{1,I}\\
\fgnore{-.25em}
&\overset{(b)}\le \nloneint{I}{\fspl\!-\!p}{\Paren{\overline{\fspl}\!-\!\bar{p}}}+\nloneint{I}{f}{p},
\fgnore{-.5em}
\end{align*}
where $(a)$ follows by the triangle inequality, and
$(b)$ follows since $\fspl$ has the same 
mass as $f$ by construction, it implies
$\|\overline{\fspl}\!-\!\bar{p}\|_{1,I}
= \!\|\bar{f}\!-\!\bar{p}\|_{1,I}
% = \|\overline{f-p}\|_{1,I}
\le \akkint{1}{I}{f-p}
\le \nloneint{I}{f}{p}$.

\fgnore{-.25em}

Since $\fspl\!-\!p\in \cP_d$, if $\nloneint{I}{q}{\bar{q}}$
is a small value $\forall q\in \cP_d$, $\fspl$ approximates $f$ nearly as well as any degree-$d$ polynomial.
Let 
\fgnore{-.25em}
\[\Delta_I(q)\ed \max_{x\in I} q(x) - \min_{x\in I} q(x)
\fgnore{-.25em}\]
be the difference between $q$'s largest and smallest values.
Note that $\barq_I(x)$ has zero mean over $I$, hence must be zero on at least one point in $I$, implying
\fgnore{-.25em}
\begin{align}
\loneint{I}{q-\barq}
&\le \Delta_{I}(q)\cdot |I|.\label{eqn:324}
\fgnore{-.25em}
\end{align}
Thus we would like
$\Delta_I(q)\cdot |I| $ 
to be small $\forall q\in \cP_d$, but 
which may not hold for the given $I$.
By additivity, we may partition $I$ and perform this adjustment over each sub-interval.
A \emph{partition} $\overline I$ of $I$ is a collection of disjoint intervals whose union is $I$.
% , where the above property holds $\forall J \in \overline I$.
Let the \emph{histogram of} $g$ \emph{over} $\overline{I}$ be 
\fgnore{-.25em}
\begin{equation}
\label{eqn:histogram}
\bar{g}_{\overline{I}}(x)
\ed
\frac{|\int_J g|}{|J|}
=\bar{g}_J(x)
\qquad x\in J\in\overline{I},
\fgnore{-.25em}
\end{equation}
We will construct a partition for which $\sum_{J\in \overline I} \Delta_J(q)\cdot |J|$ is small $ \forall q\in \cP_d$.
Further, as we don't know $f$, we will use the empirical distribution $\femp$ that
approximates the mass of $f$ over each interval of $\overline I$. 
By Lemma~\ref{lem:vc}, for any $\overline I$ with 
$k$ intervals, the expected extra error $\E\nlone{\bar{f}_{\overline{I}}}{\overline{\femp}_{\overline{I}}}=\cO(\sqrt{k/n})$.
If we want this error to be within a constant factor from the $\cO(\sqrt{(d+1)/n})$ min-max rate of $\cP_d$, 
we need to take $k=\cO(d+1)$. 
% \ignore{
% The simplest partition of an interval is into equal-length sub-intervals. 
% However, the following example, proved in Appendix~\ref{ap:unifbad}, shows
% that ...
% % that this desired property does not hold for the uniform partition.
% \begin{Example} \label{eg:unifbad}
% For $k\ge 1$, let $\overline{[-1,1]}$ partition $[-1,1]$ into $k$ equal-length intervals.
% Then for the degree-$d$ Chebyshev polynomial of the first kind, $T_d$, and for $k\ge (d+1)^3$,
% \ignore{
% \[
% \nloneint{[-1,1]}{q}{\barq_{\overline{[-1,1]}}}= \Omega(d\log d/k)\cdot \loneint{[-1,1]}{q}.
% \]
% }
% {
% \[
% \sum_{I\in \overline{[-1,1]}}\Delta_I(T_d)\cdot|I|
% = \Omega\Paren{\frac{d^2}{k}}\cdot \loneint{[-1,1]}{T_d}.
% \]
% }
% \end{Example}
% The example follows because 
% under the uniform partition $\overline{[-1,1]}$
% $\sum_{I\in \overline{[-1,1]}}\Delta_I(T_d)$
% is nearly equal to the sum of the absolute differences between the extremes of $T_d$.

\fgnore{-.25em}

We use the bound in Lemma~\ref{lem:poly} to construct a 
partition $\overline{I}$ whose widths decrease towards 
the extremes, while ensuring $k=\cO(d+1)$.
In Section~\ref{sec:proofdiff}
we show that for this
universal partition, 
$\sum_{I\in \overline{I}}\Delta_I(q)\!\cdot\! |I|$
decreases at the rate $\cO((d+1)/k)$ 
for \emph{all} $q\in \cP_d$ that we conjecture is optimal in $d$ and $k$. 

\fgnore{-.25em}

In Section~\ref{sec:actualsplit} we formally define the construction of $\fspl$ by modifying $\fpoly$ over $\overline{I}$ using $\femp$.
We show in Lemma~\ref{lem:diff2} that it suffices to select $\fpoly$ to be the polynomial estimator $\fadls$ in~\cite{jay17} or $\fsurf$ in~\cite{hao2021surf} to obtain
Theorem~\ref{thm:singlep} which shows that
$\fspl$ is a  $2$-factor approximation for $\cP_d$. 
\fgnore{-.5em}
\subsection{Polynomial Histogram Approximation}
\fgnore{-.25em}
\label{sec:proofdiff}
% \blu{
% Consider a real interval $J$ of width $|J|$ and polynomial $p\in \cP_d$. Let
% $
% \barp^J(x)
% \ed p-\bar{p}_J
% $
% be the difference between $p$ and its histogram on $J$, $\bar{p}_J$, and let 
% \[\Delta_I(p)\ed \max_{x\in I} p(x) - \min_{x\in I} p(x)\]
% be the difference between $p$'s largest and smallest values.
% Note that $\barp^J(x)$ has zero mean over the interval $J$, hence must be zero on at least one point in $J$, implying
% \begin{align*}
% \loneint{J}{\barp^J(x)}
% &\le |J|\cdot \Delta_{J}(p).\label{eqn:324old}
% \end{align*}
% }

We would first like to bound
$\Delta_I(q)$ for any $q\in \cP_d$
in terms of its $ \ell_{1} $ norm.
From the Markov Brothers' inequality~\cite{achieser1992theory}, for any $q\in\cP_d$,
\fgnore{-.25em}
\[
\Delta_{I}(q)= \cO(d+1)^2\cdot \loneint{I}{q},
\fgnore{-.25em}
\]
and is achieved by the Chebyshev polynomial of degree-$d$.
Instead, the next lemma shows that the bound can be improved for the interior of $I$. 
Its proof in Appendix~\ref{ap:poly} carefully applies Markov 
Brothers' inequality over select sub-intervals of $I$ based on the Bernstein's inequality.
For simplicity, consider $I=[-1,1]$.
\begin{Lemma}
\label{lem:poly}
For any $a\in [0,1)$ and ${q\in\cP_d}$,
% for the interval $J=[-a,a]$,
\fgnore{-.25em}
\[
\Delta_{[-a,a]}(q)
\le 
\int_{-a}^a |q'(x)|dx
\le 
\frac{28(d+1)}{\sqrt{1-a^2}}
%
% \,\lesssim\,
% \frac{d}{\sqrt{1-x^2}}
%
\loneint{[-1,1]}{q}.
\fgnore{-.25em}
\]
\end{Lemma}
We use the lemma and Equation~\ref{eqn:324} to construct a partition $ \parti{[-1,1]}{d}{k}$ of $[-1,1]$
% Example~\ref{eg:unifbad} showed that the natural choice of partitioning $[-1,1]$ into equal size intervals does not yield the necessary bound. 
% Instead, from the lemma and Equation~\eqref{eqn:324}, we construct a partition
such that $\forall J\in \parti{[-1,1]}{d}{k}$,
$\Delta_{J}(p)$ is bounded by a small value. 
Note that the lemma's bound is weaker when $a$ is close to the boundary of $(-1,1)$, hence 
the parts of $\parti{[-1,1]}{d}{k}$ decrease roughly geometrically towards the boundary, ensuring $\Delta_J(p)$ is small over each. 
The geometric partition ensures that the number of intervals is still upper bounded by $k$ as we show in Lemma~\ref{lem:jajaja}.

\fgnore{-.25em}

Consider the positive half $[0,1]$ of $[-1,1]$.
Given $\ell\ge 1$, let $m= \lceil \log_2(\ell (d+1)^2)\rceil$.
For $1\le i\le m$ define the intervals $I_i^+= [1-1/2^{i-1}, 1-1/2^i)$, that together span 
$[0,1-1/2^m)$, and let $E_{m}^+\ed [1-1/2^m,1]$ complete the partition of $[0,1]$.
Note that $ |E_{m}^+|=1/2^m \le 1 /(\ell (d+1)^2)$.
For each $1\le i\le m$ further partition $I_i^+$
into $\lceil\ell(d+1)/2^{i/4}\rceil$ intervals of equal width, and denote this partition by $\barI_i^+$.
Clearly $\barI^+\ed(\barI_1^+\upto \barI_{m}^+,E^+_{m})$ partitions $[0,1]$.

\fgnore{-.25em}

Define the mirror-image partition $\barI^-$ of $[-1,0]$, where, for example, we mirror the interval $[c,d)$ in $\barI^+$ to $(-d, -c]$.
The following lemma upper bounds the number of intervals in the combination of $\barI^-$ and $\barI^+$ and is proven in Appendix~\ref{ap:jajaja}.
% \fgnore{-.25em}
\begin{Lemma}
\fgnore{-.25em}
\label{lem:jajaja}
For any degree $d\ge 0$ and $\ell >0$,
the number of intervals in $(\barI^-,\barI^+)$ is at most $4\ell (d+1)/(2^{1/4}-1)$.
\fgnore{-.5em}
\end{Lemma}
The lemma ensures that we get the desired partition,
$\parti{[-1,1]}{d}{k}$, with $k$ intervals by setting
\fgnore{-.25em}
\begin{equation}
\label{eqn:l}
\ell \ed {k(2^{1/4}-1)}/(4(d+1)).
\fgnore{-.25em}
\end{equation}
For any interval $I=[a,b]$, we obtain $\parti{I}{d}{k}$ by a linear translation of $\parti{[-1,1]}{d}{k}$.
For example, $[c,d)\in \parti{[-1,1]}{d}{k}$ translates to 
$[a+{(b-a)(c+1)}/{2}, a+{(b-a)(d+1)}/{2})\in\parti{I}{d}{k}$. 

\fgnore{-.25em}

Recall that $\barp_{\parti{I}{d}{k}}$ denotes the histogram of $p$ on $\parti{I}{d}{k}$.
The following lemma, proven in Appendix~\ref{ap:polyhist} using Equations~\eqref{eqn:324},~\eqref{eqn:l}, and
Lemma~\ref{lem:poly},
shows that the $\ell_1$ distance of any degree-$d$ polynomial to its histogram on 
$\parti{I}{d}{k}$ is a factor $\cO((d+1)/k)$ times than the $ \ell_{1} $ norm of the polynomial.
\fgnore{-.25em}
\begin{Lemma}
\fgnore{-.25em}
\label{lem:polyhist}
Given an interval $I$, for some universal constant $c_1>1$, 
for all $p\in\cP_{d}$, and integer $k\ge 4(d+1)/(2^{1/4}-1)$,
\fgnore{-.25em}
\[
\nloneint{I}{p}{\barp_{\parti{I}{d}{k}}}\le c_1\cdot{(d+1)}\cdot\loneint{I}{p}/k.
\fgnore{-.5em}
\]
\end{Lemma}
% We obtain our estimator by adjusting
% $\fspl_{I,\fpoly, d,k}$ adjusts 
We obtain our \emph{split estimator}
$\fspl$ 
% over each $J\in \overline{I}^{d,k}$.
for a given polynomial estimator $\fpoly\in \cP_d$ 
% the \emph{split estimator} 
as 
% $\fspl \ed \fspl_{I,\fpoly,d,k}$ is defined as
\[
\fspl \ed  \fspl_{I,\fpoly,d,k}
\ed
\fpoly+
\femp_{\parti{I}{d}{k}}-\fpoly_{\parti{I}{d}{k}}
% \frac{\int_J (\femp-\fpoly)}{|J|}\ \text{for}\ x\in J\in\parti{I}{d}{k}
\]
that over each subinterval $J\in\parti{I}{d}{k}$ adds to 
$\fpoly$ a constant so that its mass over $J$ equals 
that of $\femp$. Since $\parti{I}{d}{k}$ has $k$ intervals, it follows that $\fspl\in \cP_{k,d}$.
% Denote $\fspl = \fspl_{I,\fpoly, d,k}$ for simplicity.
% such that its mass over the interval matches that of $p$. 

\fgnore{-.25em}

The next lemma (essentially) upper bounds the $\ell_1$ distance of $\fspl$ from any $p\in \cP_{d}$ that is close to $\fpoly$ in $\cA_k$ distance by the $\ell_1$ distance of $p$ from any function $f$ that is close to $\femp$ in $\cA_k$ distance.
\begin{Lemma}
\fgnore{-.25em}
\label{lem:diff}
For any interval $I$, functions $f$ and polynomials $p,\fpoly\in\cP_d$, $d\ge 0$, and $k\ge 4(d+1)/(2^{1/4}-1)$,
$\fspl=\fspl_{I,\fpoly, d,k}$ satisfies
\fgnore{-.25em}
\begin{align*}
\fgnore{-.25em}
&\nloneint{I}{\fspl}{p}\ \\
&\le\!\frac{c_1(d+1)}{k}
\nloneint{I}{\fpoly}{p}\!+\!\nloneint{I}{f}{p}\!+\!\akkint{k}{I}{\femp\!-\!f},
\fgnore{-.25em}
\end{align*}
where $c_1$ is the universal constant in Lemma~\ref{lem:polyhist}.
\end{Lemma}
\begin{Proof}
% Abbreviate $\fspl = \fspl_{I,\fpoly, d,k}$.
% $\fspcs{p}=\fpoly$ and $\fspcs{e}=\femp$. 
Consider an interval $J\in \parti{I}{d}{k}$ and let
$\overline{\fspl}, \overline{\femp}, \overline{\fpoly},\bar{p}$ respectively denote the histograms of $\fspl, \femp, \fpoly,p$ respectively over $J$. 
\fgnore{-.25em}
\begin{align*}
&\nloneint{J}{\fspl}{p}\\ &\overset{(a)}\le\!
\nloneint{J}{\fspl\!-\!p}
{\Paren{\overline{\fspl}\!-\!\bar{p}}}\!+\! \|\overline{\fspl}\!-\!\bar{p}\|_{1,J}\\
\fgnore{-.5em}
&\overset{(b)}=\!
\nloneint{J}{\fspl\!-\!p}{\Paren{\overline{\fspl}\!-\!\bar{p}}}\!+\! \!\|\overline{\femp}\!-\!\bar{p}\|_{1,J}\\
\fgnore{-.5em}
&\overset{(c)}=\!
\nloneint{J}{\fpoly\!-\!p}{\Paren{\overline{\fpoly}\!-\!\bar{p}}}\!+\! \!\|\overline{\femp}\!-\!\bar{p}\|_{1,J}\\
\fgnore{-.5em}
&\overset{(d)}\le\!
\nloneint{J}{\fpoly\!-\!p\!}{\!\Paren{\overline{\fpoly}\!\!-\!\bar{p}}}\!
+\! \!\|\overline{f}\!-\!\bar{p}\|_{1,J}\!
+\! \!\|\overline{\femp}\!-\!\bar{f}\|_{1,J},
\end{align*}
where $(a)$ and $(d)$ follow from the triangle inequality, 
$(b)$ follows since $\fspl$ has the same 
mass as $\femp$ by construction, it implies
$\|\overline{\fspl}\!-\bar{p}\|_{1,J}= \!\|\overline{\femp}\!-\bar{p}\|_{1,J}$, and
$(c)$ follows because $\fspl-\overline{\fspl}=\fpoly-\overline{\fpoly}$ since $\fspl$ and $\fpoly$ differ by a constant in each $J$.

The proof is complete by summing over $J\in \parti{I}{d}{k}$ using the fact that $\parti{I}{d}{k}$ 
has at most $k$ intervals, and
since $\fpoly\!-p\in \cP_d$ over $I$, from Lemma~\ref{lem:polyhist}, the sum 
\fgnore{-.25em}
\[\!\sum_{J\in \parti{I}{d}{k}}\!\!\!\nloneint{J}{\fpoly\!-p}{\Paren{\overline{\fpoly_J}\!-\!\bar{p}_J}}\! \le \!\!\frac{c_1(d+1)}k 
\nloneint{I}{\fpoly}{p}.
\]

% \begin{align}
%     &\overset{(a)}\le  \sum_{J\in \barI}\int\limits_{J}\Big|\fspl-p - \frac{1}{|J|}\int\limits_{J}\!\!(\fspl\!-\!p)\Big| 
% + \sum_{J\in \barI}\Big|\int\limits_{J}\!\! (\fspl\!-\!p)\Big|\\
% & \overset{(b)}= \sum_{J\in \barI}\int_{J}\Big|\fspcs{p}-p - \frac{1}{|J|}\int_{J} (\fspcs{p}-p)\Big| + \sum_{J\in \barI}\Big|\int_{J} (\fspcs{e}-p)\Big|\\
% & \overset{(c)}\le\!\! \frac{c_1(d+1)}{k}\!
% \nloneint{I}{\fspcs{p}}{p} \!\!+\! \sum_{J\in \barI}\!\Big|\!\int\limits_{J}\!\!(\fspcs{e}-f)\!\Big|\!+\!\!\sum_{J\in \barI}\Big|\!\int\limits_{J}\!\!(f-p)\!\Big|\\
% & \overset{(d)} \le \frac{c_1(d+1)}{k}
% \nloneint{I}{\fspcs{p}}{p} + \akkint{k}{I}{\fspcs{e}-f}+\nloneint{I}{f}{p},
% \end{align}

% for some constant $c_J$, $(c)$ follows from 
% Lemma~\ref{lem:polyhist} that relates the error 
% between the degree-$d$ polynomial $\fspcs{p}-p$ and its histogram on $\barI$,
\end{Proof}
% This completes the proof of Equation~\eqref{lem:diff}.
\fgnore{-1em}
\subsection{Applying the  Estimator}
\fgnore{-.25em}
\label{sec:actualsplit}
In this more technical section, we show how to use existing estimators 
in place of $\fpoly$ to achieve Theorem~\ref{thm:singlep}. 
The next lemma follows from a straightforward application of triangle inequality to Lemma~\ref{lem:diff} as shown in Appendix~\ref{ap:diff2}. It shows that given an estimate $\fpoly$ whose 
distance to $f$ is a constant multiple of $\nlone{f}{\cP_d}$
plus $\akkint{k}{I}{\femp-f}$, depending on the value of $k$, $\fspl$ 
has nearly the optimal approximation  factor of $2$ at the expense of the larger $\akkint{k}{I}{\femp -f}$.
\begin{Lemma}
% \fgnore{-.25em}
\label{lem:diff2}
Given an interval $I$, $\fpoly\in \cP_{d}$, 
such that for some constants $c', c'', \eta>0$,
\fgnore{-.25em}
\[\nloneint{I}{\fpoly}{f}\le c' \nloneint{I}{f}{\cP_{d}}+c''\akkint{k}{I}{\femp-f}+\eta,
\fgnore{-.25em}\]
and the parameter $k\ge 4(d+1)/(2^{1/4}-1)$, the estimator $\fspl=\fspl_{I,\fpoly,d,k}$ 
satisfies
\fgnore{-.25em}
\begin{align*}
\fgnore{-.25em}
\nloneint{I}{\fspl}{f}\le & 
\Paren{2+\frac{c_2(c'+1)}{k}}\nloneint{I}{f}{\cP_d}
+\frac{c_2\eta}{k}
\\
\fgnore{-.25em}
&+\Paren{1+\frac{c_2c''}{k}}
\akkint{k}{I}{\femp -f},
\fgnore{-.25em}
\end{align*}  
where $c_2 = c_1(d+1) $ 
and $c_1$ is the constant from Lemma~\ref{lem:polyhist}. 
\fgnore{-.25em}
\end{Lemma}
Prior works~\cite{jay17,hao2021surf} derive a polynomial estimator that achieves a constant factor 
approximation for $\cP_{t,d}$.
We may thus use them as $\fpoly$ in the above lemma. 
In particular, the estimator $\fadls$ in~\cite{jay17} achieves $c'=3$ and $c''=2$ and $\fsurf$ in~\cite{hao2021surf} achieves 
a $c' = c_d\ge 2$ and $c''=c_d$, where $c_d$ increases with the degree $d$ (e.g., $c'<3$ $\forall \ d\le 8$).
% Let $\omega<3$ denote the matrix multiplication exponent.
Define 
\begin{equation}
\label{eqn:eta}
\eta_d \ed \sqrt{(d+1)/n}
\end{equation}
and for any $0< \gamma<1$, let \begin{equation}
\label{eqn:k}
k(\gamma)\ed \Big \lceil {8c_1(d+1)}/\gamma\Big \rceil,
\end{equation} where $c_1$ is the constant from Lemma~\ref{lem:diff}. 
% \blu{Note that since $c_1>1, \ 0<\gamma<1$, $k(\gamma)\ge d+1$. - take care of $k$ range (re verify Lemma 7).}
We obtain the following theorem 
for $0< \gamma <1$ by
using $\fpoly = \fadls$ with $\eta = \eta_d(\gamma)$ 
in Lemma~\ref{lem:diff2}, 
and then applying Lemma~\ref{lem:vc}, all over $I=[X_{(0)}, X_{(n)}]$, i.e. the interval between the least and the largest sample.
\begin{Theorem}
% \fgnore{-.25em}
\label{thm:singlep}
Given $X^n\sim f$, for any $0<\gamma<1$, 
% \ignore{(using $\gamma$ because $\alpha$ and $\beta$ taken up in next section)}
the estimator $\fspl = \fspl_{I,  \fadls(\eta_d), d, k(\gamma)}$
for $I=[X_{(0)}, X_{(n)}]$,
achieves 
\fgnore{-.25em}
\[
\E\nlone{\fspl}{f}\le
(2+\gamma)\nlone{f}{\cP_{d}}
+\cO\Paren{\sqrt{\frac{d+1}{\gamma \cdot n}}}.
\fgnore{-.25em}
\]
\end{Theorem}
\fgnore{-.25em}
We prove the above theorem in Appendix~\ref{ap:singlep},
showing that $\fspl$ is a $2$-factor approximation for $\cP_d$.
Notice that when $\nlone{f}{\cP_d}\gg \cO\Paren{\sqrt{(d+1)/n}}$, as is the case when $n\nearrow\infty$,
Theorem~\ref{thm:singlep} gives $\fspl$ a lower $\ell_1$-distance bound to $f$ than $\fadls$. We use the above procedure 
in the main $\NADLS$ routine that we describe in the next section. 
\fgnore{-.5em}
\section{A 2-Factor Estimator for $\cP_{t,d}$}
\fgnore{-.25em}
\label{sec:multip}
In the previous section, we described an estimator that approximates a distribution
to a distance only slightly larger than twice $\nlone{f}{\cP_{1,d}}$. We now extend this result to $\cP_{t,d}$.
% \fgnore{-.5em}
% \subsection{Description}

\fgnore{-.25em}

Consider a $p^*\in \cP_{t,d}$ that achieves
$\nlone{f}{p^*}=\nlone{f}{\cP_{t,d}}$. 
If the $t$ intervals corresponding to the different polynomial pieces of $p^*$ are known, 
we may apply the routine in Section~\ref{sec:singlep} to each interval and combine the estimate
to obtain the $2$-factor approximation for $\cP_{t,d}$.

\fgnore{-.25em}

However, as these intervals are unknown, we instead use the partition returned by the $\ADLS$ routine in~\cite{jay17}. 
$\ADLS$ returns a partition 
with $\beta t$ intervals where the parameter $\beta>1$ by choice. Among these $\beta t$ intervals, $p^*\in \cP_d$
is not a degree-$d$ polynomial in at most $t$ intervals. 
Let $I$ be an interval in this partition where $p^*$ has more than one piece.
The $\ADLS$ routine has the property 
% that for any such interval, $I$, 
that there are at-least  $(\beta-1)t$ other intervals in the partition in which $p^*$ is a single-piece polynomial with a worse $\cA_{d+1}$ distance to $f$. That is, for any interval $J$ in the 
$(\beta-1)t$ interval collection,
\[\akkint{d+1}{I}{f-p^*}\le \akkint{d+1}{J}{f-p^*}+ \akkint{d+1}{J\cup I}{\femp-f}+\eta.
\]
This is used to bound the $\ell_1$ distance in these intervals. 
% as shown in Equation~\eqref{eqn:bc}.

Our main routine $\TURF$
consists of simply
applying
the transformation discussed in Section~\ref{sec:singlep} to the 
partition returned by the $\ADLS$ routine 
in~\cite{jay17}. 
Given samples $X^{n}$, the number of pieces-$t\ge 1$, 
degree-$d\ge 0$, for any $0<\alpha <1$, we first 
run the $\ADLS$ routine with input $X^{n}$, $\femp$, and parameters $t,d$, 
\fgnore{-.25em}
\begin{equation}
\label{eqn:beta}
\beta =\beta(\alpha) \ed
1+\frac{4k(\alpha)}{\alpha(d+1)},
\fgnore{-.25em}
\end{equation}
where $k(\alpha)$ is as defined in Equation~\eqref{eqn:k},
and $\eta_d = \sqrt{(d+1)/n}$.
$\ADLS$ returns a partition $\bar{I}_{\ADLS}$ of $\reals$ with $2\beta t$ intervals and a 
degree-$d$, $2\beta t$-piecewise polynomial defined over the partition. For any 
interval $I\in\bar{I}_{\ADLS}$, let $\fadls_I$ denote the degree-$d$ estimate output by $\ADLS$ over this interval.
We obtain our output estimate $\fout_{t,d, \alpha}$ by applying the routine in Section~\ref{sec:singlep} to $\fadls_I$ for 
each $I\in \barI_{\ADLS}$ with $k = k(\alpha)$ (ref. Equation~\eqref{eqn:k}). 
This is summarized below in Algorithm~\ref{alg:NADLS}.
% \fgnore{-.5em}
\begin{algorithm}[h]
\caption{$ \TURF $}
\label{alg:NADLS}
\begin{algorithmic}
\STATE {\bfseries Input:} $ X^{n} $, $ t $, $d$, $\alpha$
%\STATE $ l\gets  10/\beta$
\STATE $ k\gets \lceil {8c_1(d+1)}/{\alpha} \rceil $
\COMMENT{$c_1$ is the constant in Lemma~\ref{lem:polyhist}}
\STATE $\beta \gets 1+4k/(\alpha(d+1))$
%\STATE $c_{1}\gets\log_{2}\Paren{\sqrt2\cdot {9/10}}$
\STATE $ \eta_d \gets \sqrt{(d+1)/n}$
\STATE $ \barI_{\ADLS}, \Paren{\fadls_I, I\in \barI_{\ADLS}} \gets \ADLS(X^{n}, t,d, \beta ,\eta_d) $
% \STATE $ \barI \gets \phi $
% \STATE $ \fout_{t,d,\alpha} \gets 0$ \COMMENT{Non zero segments assigned from below loop}
% \FOR{$ I\in \barI_{\ADLS} $}
% \STATE $ \fout_{I} \gets \max\{\fout_{t,d,\alpha}, \fspl_{I,\fadls_I, d, k}\}$
% \ENDFOR
\STATE {\bfseries Output:} $\fout_{t,d,\alpha}\gets \Paren{\fspl_{I,\fadls_I, d, k}, I\in \barI_{\ADLS}}$
\end{algorithmic}
\end{algorithm}

Theorem~\ref{thm:main} shows that $\fout_{t,d,\alpha}$ is
a min-max $2 $-factor approximation for $ \cP_{t,d} $.
We have a $\cO(1/\alpha^{3/2})$ 
term in the `variance' term in Theorem~\ref{thm:main}
that reflects the $\cO(t/\alpha^{3/2})$ pieces in the output
estimate.
A small $\alpha$ corresponds to a low-bias, high-variance estimator with many pieces, and vice-versa.
Note that the $3/2$ exponent here is larger than the corresponding $1/2$ in the 
result for $\cP_d$ in Section~\ref{sec:singlep} (Theorem~\ref{thm:singlep}).
The increased exponent over $\cP_d$ is due to the unknown locations of the 
polynomial pieces of $p^*\in \cP_{t,d}$. Obtaining the exact exponent for $2$-factor approximation
for various classes may be an interesting question but beyond the scope of this paper. Let $ \omega<3 $ be the matrix multiplication constant.
As our transformation of $\fadls$ 
takes $\cO(n)$ time, the overall time complexity is the same as $\ADLS$'s near-linear $\tilde{\cO}(nd^{3+\omega})$.
\begin{Theorem}
% \fgnore{-.25em}
\label{thm:main}
Given $X^{n}\sim f $, an integer 
number of pieces $ t \ge 1 $ and degree $ d\ge 0 $, the 
parameter $ \alpha \ge 0 $,  $
\fout_{t,d,\alpha}$ is returned by $\TURF$ in $ \tilde{\cO}(n d^{3+\omega}) $ time such that 
% w.p. $\ge 2/3$,
\fgnore{-.25em}
\[
\E\nlone{\fout_{t,d,\alpha}}{f}
\le 
\Paren{2+\alpha }\nlone{f}{\cP_{t,d}}+
\cO\Paren{
\sqrt{\frac{t(d+1)}{\alpha^3 n}}}.
\fgnore{-.25em}
\]
% and $ c_{vc} $ is the constant in the VC Theorem~\cite{dev12}.
\end{Theorem}
Theorem~\ref{thm:main} is proven in Appendix~\ref{ap:main} and follows from the following lemma
via a simple application of the $\VC$ 
inequality in Lemma~\ref{lem:vc} and using Property~\ref{prop:1},~\ref{prop:2}. We prove the lemma in Appendix~\ref{sec:almostmain}.
\begin{Lemma}
\fgnore{-.25em}
\label{lem:almostmain}
Given samples $X^{n}\sim f$ for some $n\ge 1$, parameters $t\ge 1, d\ge 0$ and for $0<\alpha <1$,
$\fout_{t,d,\alpha}$ returned by $\TURF$ satisfies
\fgnore{-.25em}
\begin{align*}
&\nlone{\fout_{t,d,\alpha}}{f} \le \Paren{3+2c_1+\frac{2}{\beta-1}}\akk{2\beta t\cdot k }{\femp-f}
\\
\fgnore{-.25em}
&\phantom{lala}+
\Paren{2+\frac{4c_1(d+1)}{k}+\frac{1+k/(d+1)}{\beta-1}}\nlone{f}{\cP_{t,d}}\\
\fgnore{-.25em}
&\phantom{lala}+\Paren{\frac{c_1(d+1)}{k}+\frac{k}{(\beta-1)(d+1)}}\eta_d,
\fgnore{-.25em}
\end{align*}
where $c_1$, $k=k(\alpha)$, $\beta =\beta(\alpha)$,
are the constants in Lemma~\ref{lem:polyhist} and Equations~\eqref{eqn:k}, and~\eqref{eqn:beta} respectively, and $\eta_d \ed \sqrt{(d+1)/n}$.
\end{Lemma}
% The rest of Section~\ref{sec:multip} describes the proof of the lemma.
% \fgnore{-.25em}

\fgnore{-1em}
\section{Optimal Parameter Selection}
\fgnore{-.25em}
\label{sec:cv}
Like many other statistical learning problems, learning distributions exhibits a fundamental trade-off between bias and variance. 
In Equation~\eqref{eqn:cv} increasing the parameters $t$ and $d$ enlarges the polynomial class $\cP_{t,d}$, hence decreases the bias term $\nlone{f}{\cP_{t,d}}$ while increasing the variance term $\cO(\sqrt{t(d+1)/n})$.
As the number of samples $n$ increases, asymptotically, it is always better to opt for larger $t$ and $d$.
Yet for any given $n$, some parameters $t$ and $d$ yield the smallest error.
We consider the parameters minimizing the upper bound in Theorem~\ref{thm:main}.

\subsection{Context and Results}
\fgnore{-.25em}
\label{sec:cvcontext}
% optimal parameter values in Theorem~\ref{thm:main}.
% But as $f$ is unknown, it is infeasible in general to 
% make this selection
% % values of $t,d$ that minimizes the upper bound in Theorem~\ref{thm:main}.
% % While this optimization is infeasible for a generic $f$ 
% as it involves evaluating $\nlone{f}{\cP_{t,d}}$.
For several popular structured distributions such as unimodal, log-concave, Gaussian, and their mixtures,
low-degree polynomials, e.g. $d\le 8$, are essentially optimal~\cite{birge87, chan14, hao2021surf}.
Yet for the same classes, the range of the optimal $t$ is large, between $\Theta(1)$ and $\Theta(n^{1/3})$.
Therefore, for a given $d$, we seek the $t$ minimizing the error upper bound in Equation~\eqref{eqn:cv}.
%incurs a bias comparable to the best, as shown in Equation~\eqref{eqn:cv}.
% minimizes the upper bound on constant factor approximations for $\cP_{t,d}$.
% For any $0<\alpha<1$, $d\ge 0$, we would like to estimate $\tspcs{est}$ such that 

% Algorithm~\ref{alg:NADLS} derives an estimator  $\fout_{t,d,\alpha}$ for any $t\in\seton$.

\fgnore{-.25em}

% \cite{yatracos1985rates} derived a general cross-validation method for finding a good
% estimate for $\tspcs{est}$ for $t$.
% Applying the method to our collection of $c$-factor approximations
% $\fout_{t,d,\alpha}$ for $t\in\seton$ 
% % and using the upper bound in Theorem~\ref{thm:main},
% yields
% \begin{align*}
% \EE \lone{\fout_{\tspcs{est},d,\alpha}-f} \!\le\! \min_{t\ge 1} \Bigg(&3 c\cdot \nlone{f}{\cP_{t,d}} 
% \\
% % (\delta+ 1)  
% &+\cO\Paren{\!\sqrt{(t(d+1)\!+\!\log n)/n}}\Bigg),
% \end{align*}
% % \[
% % \E \!\|\fout_{\tspcs{est},d,\alpha}\!\!-\!f\|_1 \!\! \le  \!\min_{t\ge 1}\!
% % \Paren{\!3c\!\cdot \!\|{f}\!-\!{\cP_{t,d}}\|_1 \!+ \! \cO\Paren{\!\sqrt{t(d\!+\!1)/n}}\!},
% % \]
% % \begin{align*}
% % \E\nlone{\fout_{\tspcs{Yat},d,\alpha}}{f}\le
% % \min_{t\ge 1}&\Big(3 c\cdot \nlone{f}{\cP_{t,d}}\\
% % &+\cO(\sqrt{t(d+1)/n}\Big).
% % \end{align*}
% where $c=c(\alpha)$ is the constant in Theorem~\ref{thm:main}.

In the next subsection, we describe a parameter selection algorithm that improves this result
for the estimators we considered in the previous section. 
Following nearly identical steps as in the derivation of Theorem~\ref{thm:main} from Lemma~\ref{lem:almostmain}, and using
the probabilistic version of the VC Lemma~\ref{lem:vc} (see \cite{dev12}), it may be shown that with high probability 
$\fout_{t,d,\alpha}$ is a
$c$-factor approximation for
$\cP_{t,d}$. Namely, for any $\delta\ge 0$,
\begin{equation}
\label{eqn:hp}
\!\|\fout_{t,d,\alpha}-\!f\|_1 \! \le 
\!c\cdot \|{f}\!-\!{\cP_{t,d}}\|_1 \!+  \cO\Paren{\!\sqrt{(t(d\!+\!1)+\log 1/\delta)/n}}, 
\end{equation}
where $c=c(\alpha)$ is a function of the chosen $\alpha$.
% where $c=c(\alpha)$ is the constant in Theorem~\ref{thm:main}.
% Given the sequence ... of estimators, 
We use the estimates $\fout_{t,d,\alpha}$ to find an estimate $\tspcs{est}$ such that $\fout_{\tspcs{est},d,\alpha}$ has an error comparable to the $c$-factor approximation for $\cP_{t,d}$ with the best $t$.
% resulting in a lower error for large $n$.

% at the expense of a larger variance term.
% proved in Appendix~\ref{ap:middle}
% Parameterized by a given $\delta>2$, our estimator $\tspcs{est}_{\delta}$, is built using the sequence of estimates $\fout_{t,d,\alpha}$, $t\in \{1\upto n\}$
% for a given $d,\alpha$
% as we describe in Section~\ref{sec:cvconstruct}.

% and shows that depending on the value of $\delta$, $\fspcs{out}_{\tspcs{est}_{\delta},d}$ has a bias comparable to the best pre-selected value of $t$ at the expense of a larger variance term.
\begin{Theorem}
\label{thm:middle}
Given $n\in 2^{\cN}$, $d\ge 0$, $0<\alpha<1$,
$c$-factor estimates for $\cP_{t,d}$ in high probability (see Equation~\eqref{eqn:hp}),
$\{\fout_{t,d,\alpha} :1\le t\le n\}$,
for any $0<\beta<1$, we find the estimate $\tspcs{est}$ such that
w.p. $\ge 1-\delta \cdot \log n$,
% selecting $\delta=2+1/\beta$
\fgnore{-.25em}
\begin{align*}
\lone{\fout_{\tspcs{est},d,\alpha}-f} \!\le\! &\min_{t\ge 1} \Bigg(\Paren{1+\beta} \cdot c\cdot \nlone{f}{\cP_{t,d}} 
\\
% (\delta+ 1)  
&+\cO\Paren{\!\sqrt{(t(d+1)\!+\!\log 1/\delta)/(\beta^2 n)}}\Bigg).
\fgnore{-.25em}
\end{align*}
\end{Theorem}
The proof, provided in Appendix~\ref{ap:middle}, exploits the fact that the bias term of $\fout_{t,d,\alpha}$ is at most $c\cdot \nlone{f}{\cP_{t,d}}$, which decreases with  $t$, and the variance term upper bounded by $\cO(\sqrt{(t(d+1)+\log1/\delta)/n})$, increasing with $t$.

% We use these two properties to derive a method that works in a general metric space $(\cM,d)$.

% As we show in~\cite{}, this method also has 
% applications in general robust learning,
% where we show that it sidesteps the need for apriori knowing the \emph{corruption level} that existing works assume,
% and may be of independent interest.
% \fgnore{-.25em}
% Finally, observe 
% from Theorem~\ref{thm:middle} that for a large value of $n$, when $\delta$ is chosen to be sufficiently large, $\nlone{\fspcs{out}_{\tspcs{est}_{\delta},d}}{f}
% \approx \min_{t\ge 1} \nlone{\fspcs{out}_{t,d,\alpha}}{f}$. Thus we achieve a min-max estimator with nearly the optimal bias across $t$.
\fgnore{-.25em}
\subsection{Construction}
\fgnore{-.25em}
\label{sec:cvconstruct}
We use the following algorithm derived in~\cite{JORcomm}. Consider the set $\cV= \{v_1, v_2,\upto v_k\}\subseteq \cM$, an \emph{unknown} target 
$v\in \cM$, an \emph{unknown} non-increasing sequence $b_i$, and a \emph{known} non-decreasing sequence 
$c_i$ such that $\dist{v_i}{v}\le b_i+c_i \ \forall i$.

\fgnore{-.25em}

First consider selecting for a given $1\le i<j\le k$, the point among $v_i, v_j$ 
that is closer to $v$.
Suppose for some constant $\gamma>0$,
$\dist{v_i}{v_j}\le \gamma c_j$. Then from the triangle inequality, $d(v_i,v)\le d(v_j,v)+d(v_i,v_j)\le 
b_j+c_j+\gamma c_j\le b_j+(1+\gamma)c_j$.
% Therefore in this case we may select $v_i$.
On the other hand if $\dist{v_i}{v_j}> \gamma c_j$, since $b_j\le b_i$ (as $j>i$),
$d(v_j,v)\le b_j+c_j\le  b_i+c_j\le 
b_i+\dist{v_i}{v_j}/\gamma$. 

\fgnore{-.25em}

Therefore if we set $\gamma$ to be sufficiently 
large and select $v'_{\gamma}=v_i$ if $\dist{v_i}{v_j}\le \gamma c_j$, 
and otherwise set $v'_{\gamma}=v_j$, we roughly obtain $d(v'_{\gamma},v)\lesssim b_i+(1+\gamma)c_j$. 
We now generalize this approach to selecting between all points in $\cV$. 
Let $i_\gamma$ be the smallest index in $ \sett{k}$ 
such that $\forall i_\gamma<i\le k$, $\dist{v_{i_\gamma}}{v_j}\le \gamma c_i$. 
Lemma~\ref{lem:cvini} shows the favorable properties of $v_{i_\gamma}$, 
for example, that for a sufficiently large $\gamma$, $\dist{v_{i_\gamma}}{v}$
is comparable to $\min_{i\in\{1\upto k\}}(b_i+\lambda c_i)$ when $b_i\gg c_i$. 
The proof may be found in Appendix~\ref{ap:cvini}. 
\begin{Lemma}
\label{lem:cvini}
% \fgnore{-.25em}
Given a set $\cV = \{v_1, v_2,\upto v_k\}$ in a metric space $(\cM,d)$, 
a sequence $0\le c_1\le c_2\ldots\le c_k$,
and $\gamma>2$, 
let $1\le i_\gamma\le k$ be the smallest index such that for all
$i_\gamma<i\le k$,
$\dist{v_i}{v_{i_\gamma}}\le \gamma c_i$.
Then for all sequences $b_1\ge b_2\ge\ldots\ge b_k\ge0$ such that for all $i$,
$\dist{v_i}{v}\le b_i +  c_i$, 
\fgnore{-.25em}
\[
\dist{v_{i_\gamma}}{v}
\le
\min_{j\in \sett{k}}
\Paren{
\Paren{1+\frac{2}{\gamma -2}}\cdot b_{j}+
(\gamma+1) c_j}.
\fgnore{-.25em}
\]
\end{Lemma}
The set of real integrable functions with $\tvv$
distance forms a metric space.
For simplicity for the given $d\ge 0$, $0<\alpha<1$, and $n$, denote $\fout_{t}\ed \fout_{t,d,\alpha}$. Assume $n$ is a power of $2$ and let $\cV = \{\fout_{1}, \fout_{2}, \fout_{4}\upto \fout_{n}\}$.
Suppose for constants $c', c''>0$, and 
a chosen $0<\delta<1$,
$\sqrt{c't+c''\log1/\delta}$ is $\fout_t$'s variance term.
For any chosen $0<\beta<1$, we obtain $\tspcs{est}_{\beta}$ by applying the above method with
$\cV$ and the $c_i$s corresponding to $\fout_{t}$
as $\sqrt{c't+c''\log1/\delta}$.
That is, $\tspcs{est}_{\beta}$ is the smallest $t\in \cI = \{1,2,4\upto n\}$ such that $\forall j\in \cI: j\ge t $,
$\dist{\fout_{t}}{\fout_{j}}
\le \gamma \sqrt{c'j+c''\log1/\delta}$ (where we select $\gamma=\gamma(\beta)=2+2/\beta$).
In Section~\ref{sec:experi}, we experimentally evaluate the $\NADLS$ estimator and the cross-validation technique.
\fgnore{-1em}
\section{Experiments}
\fgnore{-.25em}
\label{sec:experi}
\begin{figure*}[!ht]
\centering
\subfigure{\includegraphics[scale=0.34]{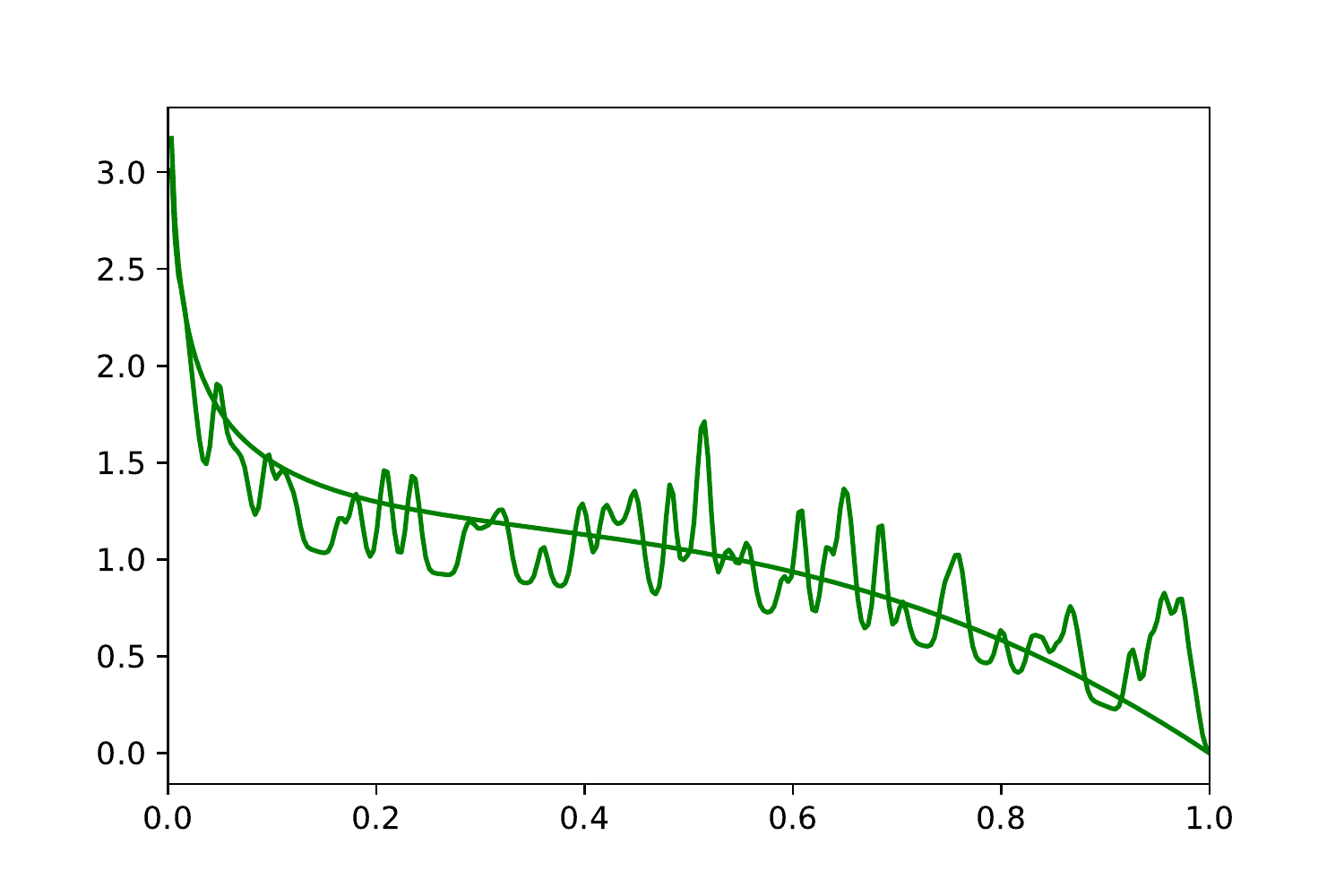}\label{aaa}}\quad
\subfigure{\includegraphics[scale=0.34]{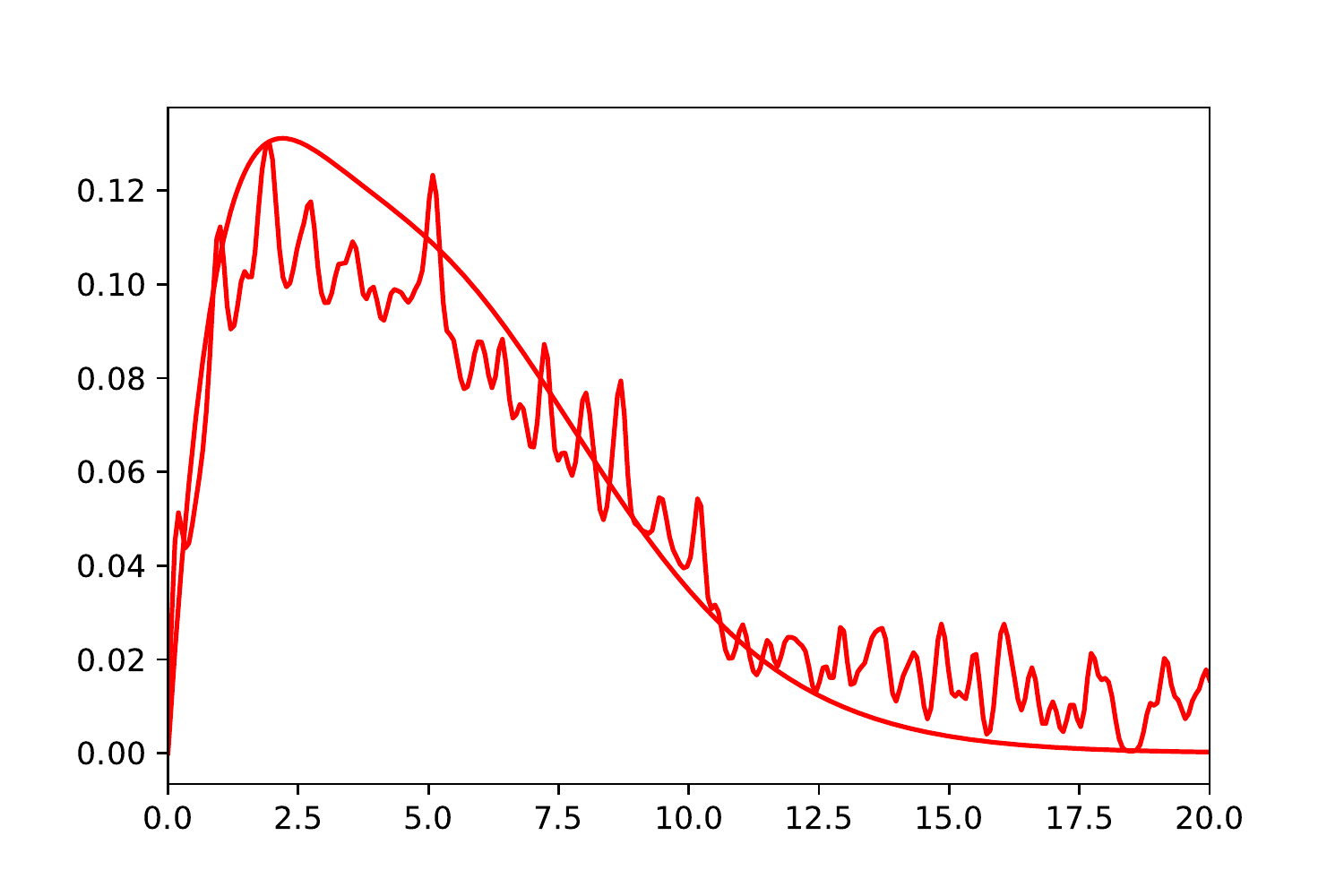}\label{bbb}}\quad
\subfigure{\includegraphics[scale=0.34]{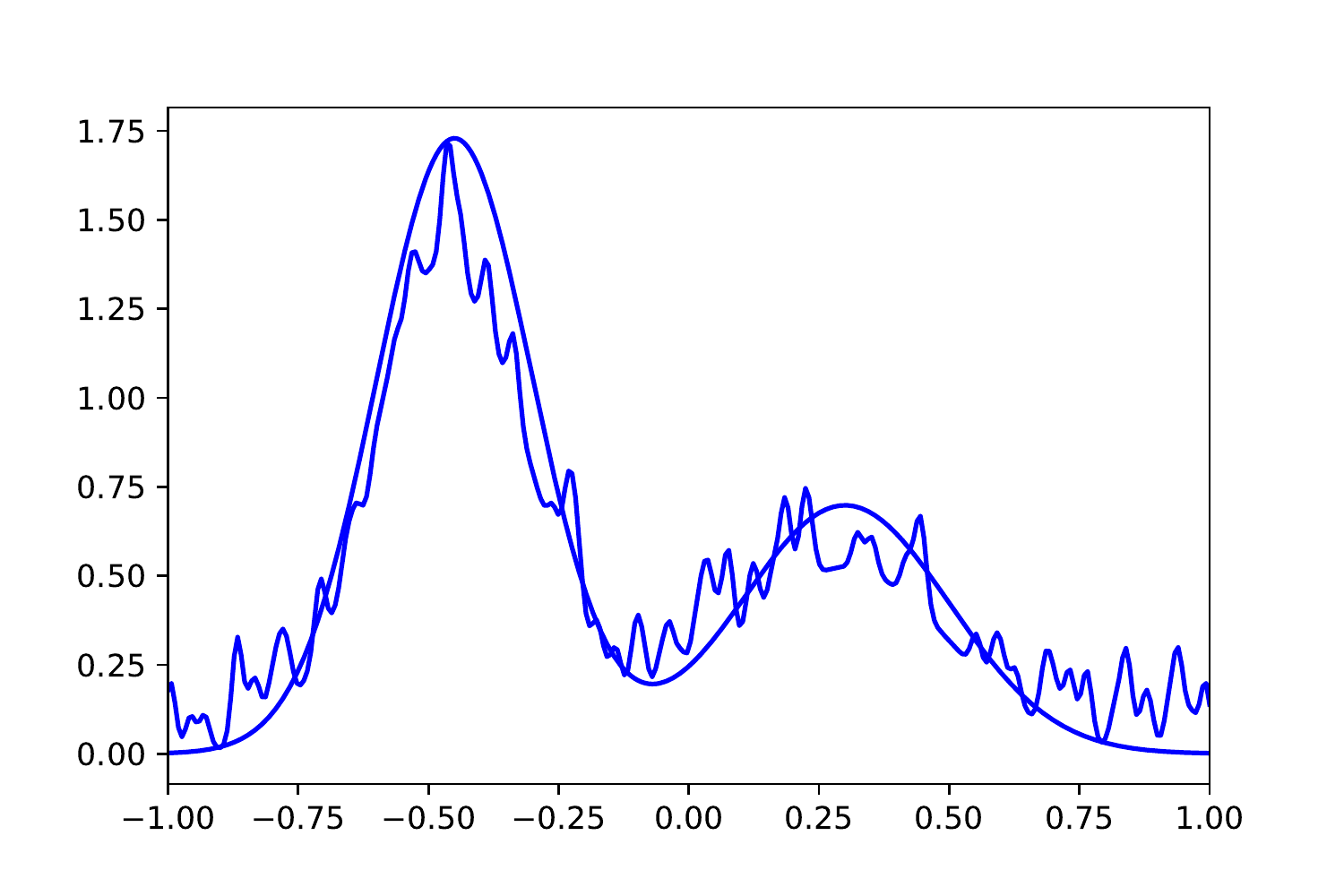}\label{ccc}}
\vspace{-1em}
\caption{The Beta, Gamma and Gaussian mixtures, respectively. The smooth and coarse plots in each sub-figure correspond to the noise-free and noisy cases, respectively.}
\vspace{-1.25em}
\label{plot:compare0}
\end{figure*}
\begin{figure*}[!ht]
\centering
\subfigure{\includegraphics[scale=1.1]{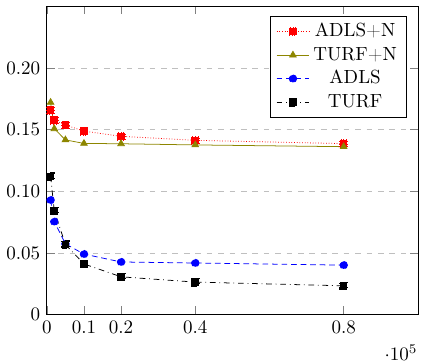}\label{aa}}\qquad
\subfigure{\includegraphics[scale=1.1]{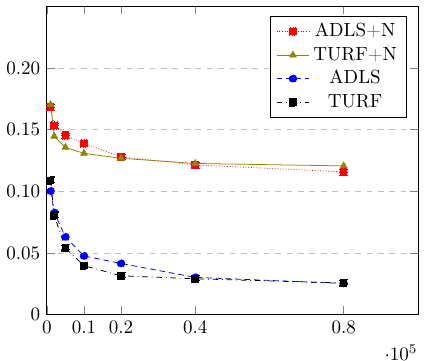}\label{bb}}\qquad
\subfigure{\includegraphics[scale=1.1]{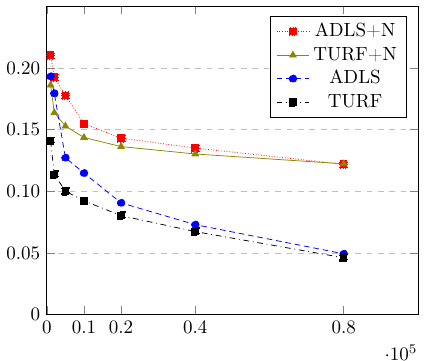}\label{cc}}
\vspace{-1em}
\caption{$\ell_{1} $ error versus number of samples on the Beta, Gamma, and Gaussian mixtures respectively in Figure~\ref{plot:compare0}
for $d=1$.}
\vspace{-1.25em}
\label{plot:compare}
\end{figure*}
\begin{figure*}[!ht]
\centering
\subfigure{\includegraphics[scale=1.1]{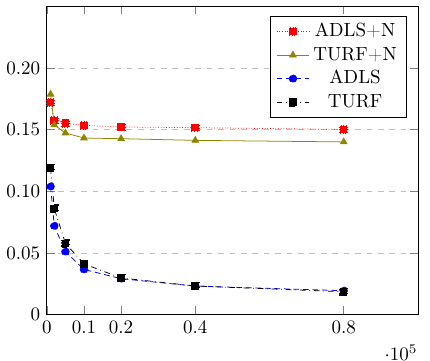}\label{aaa}}\qquad
\subfigure{\includegraphics[scale=1.1]{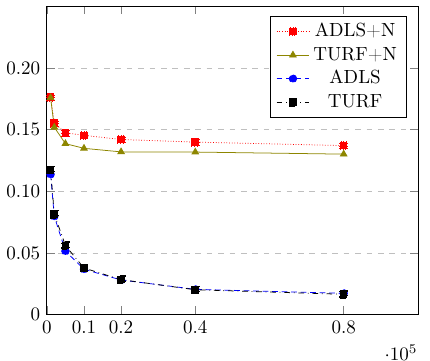}\label{bbb}}\qquad
\subfigure{\includegraphics[scale=1.1]{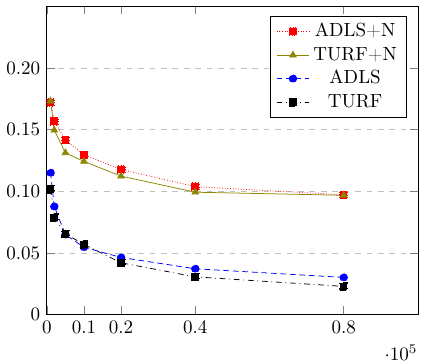}\label{ccc}}
\vspace{-1em}
\caption{$\ell_{1} $ error versus number of samples on the Beta, Gamma, and Gaussian mixtures respectively in Figure~\ref{plot:compare0} for $d=2$.}
\vspace{-1.25em}
\label{plot:compare1}
\end{figure*}
Direct comparison of $\TURF$ and $\ADLS$ for a given $t,d$ is not straightforward as $\TURF$ outputs polynomials consisting of more pieces. 
To compare the algorithms more equitably, we apply the cross-validation technique in Section~\ref{sec:cv} to select the best $t$ for each. 
The cross validation parameter $\delta$ is chosen to reflect the actual number of pieces output by $\ADLS$ and $\TURF$.
Note that while $\SURF$~\cite{hao2021surf} is another piecewise polynomial based estimation method, it has an implicit method to cross-validate $t$, unlike 
$\ADLS$ and $\TURF$. As comparisons against $\SURF$ may only reflect the relative strengths of the cross validation methods 
and not that of the underlying estimation procedure, we defer them to Appendix~\ref{sec:appen6}.
All experiments compare the $\ell_1$ error, run for $n$ between 1,000 and 80,000, and averaged over 50 runs.
For $ \ADLS $ we use the code provided in~\cite{jay17}, and for $\TURF$ we use the algorithm in Section~\ref{sec:multip}.\looseness-1
% splits the polynomial pieces of $\ADLS$'s estimate along select sub-intervals to match the empirical mass.
% In all experiments $ \TURF $ is obtained by 
% modifying $\ADLS$'s estimate. 

\fgnore{-.25em}

The experiments consider the structured distributions addressed in~\cite{jay17}, namely mixtures of Beta: $.4\text{B}(.8, 4)+.6\text{B}(2, 2) $,
Gamma: $.7\Gamma(2, 2)+.3\Gamma(7.5, 1) $, and Gaussians: .65$\cN$(-.45,$.15^2$)+.35$\cN$(.3,$.2^2$) as shown in Figure~\ref{plot:compare0}.
Figure~\ref{plot:compare} considers approximation relative to $\cP_{t,1}$.
The blue-dashed and the black-dot-dash plots show that $\TURF$ modestly outperforms $\ADLS$. It is especially significant for the Beta distribution as $\text{B}(.8,4)$ 
has a large second derivative near $0$, and approximating it may require many degree-1 pieces localized to that region. For this lower width region, the $\cA_1$ distance
may be too small to warrant many pieces in $\ADLS$, unlike in $\TURF$
that forms intervals guided by shape constraints e.g., based on Lemma~\ref{lem:poly}.
% The red....

\fgnore{-.25em}

We perturb these distribution mixtures to increase their bias.
For a given $ k>0 $, select $\bar{\mu }_{k}\ed (\mu_{1}\upto \mu_{k}) $
by independently choosing $ \mu_{i}, \ i\in \sett{k} $
uniformly from the effective support of $ f $ (we remove $5\%$ tail mass on either side).
At each of these locations, apply a Gaussian noise of magnitude $0.25/k$ with standard deviation $\sigma=c_2/k$, for some constant $c_2>0$ that is chosen to scale with the effective support width of $f$. That is,
\fgnore{-.25em}
\[ 
f_{\bar{\mu}} \ed \frac34 \cdot f + \frac14 \cdot \sum_{i=1}^{k} \frac1k \cdot \cN\Paren{\mu_{i}, \frac{c_2^2}{k^2}}.
\fgnore{-.25em}\] 
We choose $k=100$ and $c_2=0.05,1,0.1$ for the Beta, Gamma and Gaussian mixtures respectively, to yield the distributions shown in Figure~\ref{plot:compare0}.
The red-dotted and olive-solid plots in Figure~\ref{plot:compare} compares $ \ADLS $
and $ \TURF $ on these distributions. While the overall errors are larger due to the added noise, $ \TURF $ outperforms $\ADLS$ on nearly all distributions. A consistent trend across our experiments is that for large $n$, the performance gap between $\ADLS$ and $\TURF$ decreases. This may be explained by the fact that as $n$ increases, the value of $t$ output by the cross-validation method also increases, reducing the bias under both $\ADLS$ and $\TURF$. However, the reduction in $\ADLS$’s bias is more significant due to its larger approximation factor compared to $\TURF$, resulting in the smaller gap.

Figure~\ref{plot:compare1} repeats the same experiments for $d=2$.
Increasing the degree leads to lower errors on both $\ADLS$ and $\TURF$ in the non-noisy case. However, the larger bias in the noisy case reveals the improved performance of $\TURF$.

% \section{Acknowledgements}
% We are grateful to the National Science Foundation (NSF) for supporting this work through grants CIF-1564355 and CIF-1619448.

%\input{experi.tex}

\bibliography{ref}

\begin{thebibliography}{32}
\providecommand{\natexlab}[1]{#1}
\providecommand{\url}[1]{\texttt{#1}}
\expandafter\ifx\csname urlstyle\endcsname\relax
  \providecommand{\doi}[1]{doi: #1}\else
  \providecommand{\doi}{doi: \begingroup \urlstyle{rm}\Url}\fi

\bibitem[Acharya et~al.(2014)Acharya, Jafarpour, Orlitsky, and
  Suresh]{acharya2014near}
Acharya, J., Jafarpour, A., Orlitsky, A., and Suresh, A.~T.
\newblock Near-optimal-sample estimators for spherical gaussian mixtures.
\newblock \emph{arXiv preprint arXiv:1402.4746}, 2014.

\bibitem[Acharya et~al.(2017)Acharya, Diakonikolas, Li, and Schmidt]{jay17}
Acharya, J., Diakonikolas, I., Li, J., and Schmidt, L.
\newblock Sample-optimal density estimation in nearly-linear time.
\newblock In \emph{Proceedings of the Twenty-Eighth Annual ACM-SIAM Symposium
  on Discrete Algorithms}, pp.\  1278--1289. SIAM, 2017.

\bibitem[Achieser(1992)]{achieser1992theory}
Achieser, N.
\newblock \emph{Theory of Approximation}.
\newblock Dover books on advanced mathematics. Dover Publications, 1992.
\newblock ISBN 9780486671291.

\bibitem[Ashtiani et~al.(2018)Ashtiani, Ben-David, Harvey, Liaw, Mehrabian, and
  Plan]{ash18}
Ashtiani, H., Ben-David, S., Harvey, N.~J., Liaw, C., Mehrabian, A., and Plan,
  Y.
\newblock Nearly tight sample complexity bounds for learning mixtures of
  gaussians via sample compression schemes.
\newblock In \emph{Proceedings of the 32nd International Conference on Neural
  Information Processing Systems}, pp.\  3416--3425, 2018.

\bibitem[Birg{\'e}(1987)]{birge87}
Birg{\'e}, L.
\newblock Estimating a density under order restrictions: Nonasymptotic minimax
  risk.
\newblock \emph{The Annals of Statistics}, pp.\  995--1012, 1987.

\bibitem[Bithell(1990)]{bithell1990application}
Bithell, J.~F.
\newblock An application of density estimation to geographical epidemiology.
\newblock \emph{Statistics in medicine}, 9\penalty0 (6):\penalty0 691--701,
  1990.

\bibitem[Bousquet et~al.(2019)Bousquet, Kane, and Moran]{bousquet19}
Bousquet, O., Kane, D., and Moran, S.
\newblock The optimal approximation factor in density estimation.
\newblock \emph{arXiv preprint arXiv:1902.05876}, 2019.

\bibitem[Bousquet et~al.(2021)Bousquet, Braverman, Efremenko, Kol, and
  Moran]{bousquet2021statistically}
Bousquet, O., Braverman, M., Efremenko, K., Kol, G., and Moran, S.
\newblock Statistically near-optimal hypothesis selection.
\newblock \emph{arXiv preprint arXiv:2108.07880}, 2021.

\bibitem[Chan et~al.(2014)Chan, Diakonikolas, Servedio, and Sun]{chan14}
Chan, S.-O., Diakonikolas, I., Servedio, R.~A., and Sun, X.
\newblock Efficient density estimation via piecewise polynomial approximation.
\newblock In \emph{Proceedings of the forty-sixth annual ACM symposium on
  Theory of computing}, pp.\  604--613. ACM, 2014.

\bibitem[Cohen et~al.(2020)Cohen, Kontorovich, and Wolfer]{cohen2020learning}
Cohen, D., Kontorovich, A., and Wolfer, G.
\newblock Learning discrete distributions with infinite support.
\newblock \emph{Advances in Neural Information Processing Systems},
  33:\penalty0 3942--3951, 2020.

\bibitem[Devroye \& Gyorfi(1990)Devroye and Gyorfi]{devroye1990no}
Devroye, L. and Gyorfi, L.
\newblock No empirical probability measure can converge in the total variation
  sense for all distributions.
\newblock \emph{The Annals of Statistics}, pp.\  1496--1499, 1990.

\bibitem[Devroye \& Lugosi(2012)Devroye and Lugosi]{dev12}
Devroye, L. and Lugosi, G.
\newblock \emph{Combinatorial methods in density estimation}.
\newblock Springer Science \& Business Media, 2012.

\bibitem[Diakonikolas(2016)]{diakonikolas2016learning}
Diakonikolas, I.
\newblock Learning structured distributions.
\newblock \emph{Handbook of Big Data}, pp.\  267, 2016.

\bibitem[Gerber(2014)]{gerber2014predicting}
Gerber, M.~S.
\newblock Predicting crime using twitter and kernel density estimation.
\newblock \emph{Decision Support Systems}, 61:\penalty0 115--125, 2014.

\bibitem[Givens \& Hoeting(2012)Givens and Hoeting]{givens2012computational}
Givens, G.~H. and Hoeting, J.~A.
\newblock \emph{Computational statistics}, volume 703.
\newblock John Wiley \& Sons, 2012.

\bibitem[Goodfellow et~al.(2014)Goodfellow, Pouget-Abadie, Mirza, Xu,
  Warde-Farley, Ozair, Courville, and Bengio]{goodfellow2014generative}
Goodfellow, I., Pouget-Abadie, J., Mirza, M., Xu, B., Warde-Farley, D., Ozair,
  S., Courville, A., and Bengio, Y.
\newblock Generative adversarial nets.
\newblock \emph{Advances in neural information processing systems}, 27, 2014.

\bibitem[Han et~al.(2015)Han, Jiao, and Weissman]{han2015minimax}
Han, Y., Jiao, J., and Weissman, T.
\newblock Minimax estimation of discrete distributions under $\ell_1$ loss.
\newblock \emph{IEEE Transactions on Information Theory}, 61\penalty0
  (11):\penalty0 6343--6354, 2015.

\bibitem[Hao \& Orlitsky(2019)Hao and Orlitsky]{hao2020unified}
Hao, Y. and Orlitsky, A.
\newblock Unified sample-optimal property estimation in near-linear time.
\newblock \emph{Advances in Neural Information Processing Systems}, 32, 2019.

\bibitem[Hao et~al.(2020)Hao, Jain, Orlitsky, and Ravindrakumar]{hao2021surf}
Hao, Y., Jain, A., Orlitsky, A., and Ravindrakumar, V.
\newblock Surf: A simple, universal, robust, fast distribution learning
  algorithm.
\newblock \emph{Advances in Neural Information Processing Systems},
  33:\penalty0 10881--10890, 2020.

\bibitem[Huber(1992)]{huber1992robust}
Huber, P.~J.
\newblock Robust estimation of a location parameter.
\newblock In \emph{Breakthroughs in statistics}, pp.\  492--518. Springer,
  1992.

\bibitem[Jain et~al.(2022)Jain, Orlitsky, and Ravindrakumar]{JORcomm}
Jain, A., Orlitsky, A., and Ravindrakumar, V.
\newblock Robust estimation algorithms don't need to know the corruption level.
\newblock \emph{arXiv preprint arXiv:2202.05453}, 2022.

\bibitem[Kamath et~al.(2015)Kamath, Orlitsky, Pichapati, and
  Suresh]{kamath2015learning}
Kamath, S., Orlitsky, A., Pichapati, D., and Suresh, A.~T.
\newblock On learning distributions from their samples.
\newblock In \emph{Conference on Learning Theory}, pp.\  1066--1100. PMLR,
  2015.

\bibitem[Pearson(1895)]{pea95}
Pearson, K.
\newblock X. contributions to the mathematical theory of evolution.—ii. skew
  variation in homogeneous material.
\newblock \emph{Philosophical Transactions of the Royal Society of
  London.(A.)}, \penalty0 (186):\penalty0 343--414, 1895.

\bibitem[Pimentel et~al.(2014)Pimentel, Clifton, Clifton, and
  Tarassenko]{pimentel2014review}
Pimentel, M.~A., Clifton, D.~A., Clifton, L., and Tarassenko, L.
\newblock A review of novelty detection.
\newblock \emph{Signal Processing}, 99:\penalty0 215--249, 2014.

\bibitem[Rahman et~al.(2002)Rahman, Schmeisser, et~al.]{rahman2002analytic}
Rahman, Q.~I., Schmeisser, G., et~al.
\newblock \emph{Analytic theory of polynomials}.
\newblock Number~26 in London Mathematical Society monographs. Clarendon Press,
  2002.

\bibitem[Scott(2012)]{sco12}
Scott, D.~W.
\newblock Multivariate density estimation and visualization.
\newblock In \emph{Handbook of computational statistics}, pp.\  549--569.
  Springer, 2012.

\bibitem[Silverman(1986)]{sil86}
Silverman, B.~W.
\newblock \emph{Density Estimation for Statistics and Data Analysis},
  volume~26.
\newblock CRC Press, 1986.

\bibitem[Tukey(1977)]{tukey1977exploratory}
Tukey, J.~W.
\newblock Exploratory data analysis.
\newblock \emph{Addison-Wesley Series in Behavioral Science: Quantitative
  Methods}, 1977.

\bibitem[Vapnik(1999)]{vapnik1999overview}
Vapnik, V.~N.
\newblock An overview of statistical learning theory.
\newblock \emph{IEEE transactions on neural networks}, 10\penalty0
  (5):\penalty0 988--999, 1999.

\bibitem[Wolfowitz(1957)]{wolfowitz1957minimum}
Wolfowitz, J.
\newblock The minimum distance method.
\newblock \emph{The Annals of Mathematical Statistics}, pp.\  75--88, 1957.

\bibitem[Yatracos(1985)]{yatracos1985rates}
Yatracos, Y.~G.
\newblock Rates of convergence of minimum distance estimators and kolmogorov's
  entropy.
\newblock \emph{The Annals of Statistics}, pp.\  768--774, 1985.

\bibitem[Zambom \& Ronaldo(2013)Zambom and Ronaldo]{zambom2013review}
Zambom, A.~Z. and Ronaldo, D.
\newblock A review of kernel density estimation with applications to
  econometrics.
\newblock \emph{International Econometric Review}, 5\penalty0 (1):\penalty0
  20--42, 2013.

\end{thebibliography}
\bibliographystyle{icml2020}

\onecolumn
\appendix
\section{Proofs for Section~\ref{sec:introduction}}
\label{sec:appen}

\subsection{Proof of Lemma~\ref{lem:example2}}
\label{pf:example2}
\begin{Proof}
\begin{figure}[h]
\centering
\includegraphics[scale=1.1]{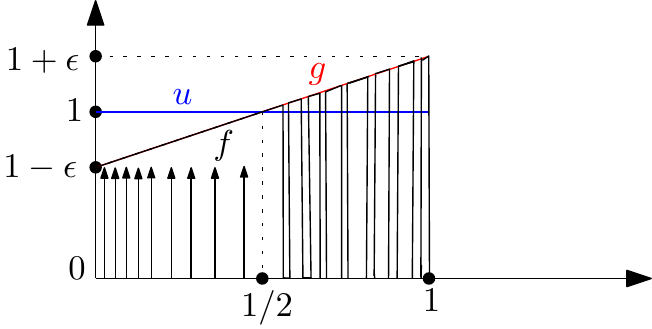}
\caption{$f$ that is indistinguishable from $u$ in the proof of Lemma~\ref{lem:example2}}
\label{fig:example2}
\end{figure}
% The following definition is helpful in the following proof.
% \begin{Definition}

% \end{Definition}

Note that for any $t_1\le t$ and $d_1\le d$, $\cP_{t_1,d_1} \subseteq \cP_{t,d}$.
Therefore
$c_{t,d}$ increases with $t,d$.
\cite{chan14} showed that $c_{2,0}\ge 2$. We show below that that $c_{1,1}\ge 2$. 
Together they imply $c_{t,d}\ge 2$ when $t\ge 2$ or $d\ge 1$.

Let $u$ be the uniform distribution on $[0,1]$. Fix an $\epsilon>0$ and consider the distribution
$g(x)=1-\epsilon+2\epsilon x$. Note that $g\in \cP_{1,1}$.
For a fixed $k\ge 1$, we construct two random distributions $f_k$ and $f_k'$
that are essentially indistinguishable
using $n$ samples as $k$ is made large,
and such that all functions have an $\ell_1$ distance to either $f_k$ or $f_k'$ that is at least twice as far as their respective $\ell_1$ approximations in $\cP_{1,1}$.

To construct $f_k$ we perturb $g$ separately on the left, $[0,1/2]$, and right, $(1/2,1]$, halves of $[0,1]$.

For the left half, we use the discrete sub-distribution $h_k$ that assigns a mass 
$\epsilon/4\cdot 1/k$ to $k$ values drawn according to the distribution $h(x)=4-8x$ over $[0,1/2]$.
Then let
\[
f_k(x) = g(x)+h_k(x),\text{\quad for\quad}  x\in [0,1/2].
\]
Thus $f_k$ consists of discrete atoms added to $g$ on $[0,1/2]$.
% By a birthday paradox type argument, it is easy to show that for large $k$, $f_k$ cannot be distinguished from $u$ using $n$ samples on $[0,1/2]$.  

For the right half, assuming Wolog that $k$ is even, first partition $(1/2,1]$ into $k/2$ intervals of width $1/k$ by letting
$I_i \ed (1/2+(i-1)/k,1/2+i/k]$ for $i\in \sett{k/2}$.
Let $|I|$ denote the width of interval $I$. 
%To construct $f_k$ on $(1/2,1]$ we choose 
For each $i\in\sett{k/2}$, select a random circular sub-interval $J_i\subseteq I_i$ of width 
\[
w_i = \frac{1}{1+2\epsilon i/k}\cdot |I_i|
\] 
as follows: Suppose $I_i=[a_i,b_i]$ for simplicity.
Choose a point $x_i$ uniformly at random in $I_i$ and define 
\[J_i \ed [
a_i, a_i+
\max\{0,x_i+w_i-b_i\}
]\cup (x_i,\min\{x_i+w_i,b_i\}] .\] 
Let $f_k$ be $g$ over $J_i$ and $0$ over $I_i\setminus J_i$, hence
% Specifically, for each $i\in\sett{k/2}$, we set $f_k = g$ on a random sub-interval, $J_i$, of $I_i$ of width \[|J_i|=\frac{1}{1-\epsilon+2\epsilon \Paren{1/2+i/k}}\cdot |I_i|\] 
% chosen uniformly at random from $I_i$.
% Assign $f_k = 0$ on its complement $K_i \ed I_i\setminus J_i$.
% that is of width $|K_i|=(2\epsilon x-\epsilon)/(1-\epsilon+2\epsilon x)\cdot |I_i|$.
as illustrated in Figure~\ref{fig:example2}, for $x\in (1/2,1]$,
\[
f_k(x)\ed 
\begin{cases}
g(x) & x \in J_i,\quad i\in \sett{k/2},\\
0 & x \in I_i\setminus J_i,\quad i \in \sett{k/2}.
\end{cases}
\]
It is easy to show that on any sub-interval of $I_i$, 
the area of $f_k$ is within that of $u$ up to an additive $\cO(\epsilon/k)$. 
% Let $f\ed \lim_{k\rightarrow \infty} f_k$.

% Note that on $[0,1/2]$, $g\in \cP_{1,1}$ is the best linear approximation to $f_k$ as the atoms in $h_k$ cannot be approximated by any continuous distribution. This also follows on the RHS $(1/2,1]$, 
% Since
% $|J_i|\ge |I_i|/2$ $\forall i \in \sett{k/2}$ for any $\epsilon\le 1$ and as the discrete component of $f_k(x)$ on $[0,1]$ cannot be approximated with any continuous function, it follows that $\nlone{f_k}{\cP_{1,1}}$ is achieved by $g\in \cP_{1,1}$.
% Let $f\ed \lim_{k\rightarrow \infty} f_k$.
Construct $f_k'$ via the same method for $f_k$ but mirrored along $1/2$, with $g'=1+\epsilon-2\epsilon x$, adding atoms to $(1/2,1]$ and alternating between $g'$ and $0$ on $[0,1/2]$ as described in the construction of $f_k$.

By a birthday paradox type argument, for any $\delta\ge 0$, it is easy to see that the distributions $f_k$, $f_k'$ are 
indistinguishable from $u$
with probability $\ge 1-\delta$ using any finitely many $n$ samples (by choosing an appropriately large $k=k(\delta,n)$). Thus  w.p. $\ge 1-2\delta$, 
the estimate $\fest = \fest_{X^n}$ is identical under both $f_k$ and $f_k'$.
% Wolog suppose that with high probability $\fest$ point-wise converges to some function as $n\nearrow \infty$ for $X^n\sim f \text{ or }f_0$. 
Therefore any estimator $\fest$ suffers a factor
\[
c\ge 
(1-2\delta) \cdot \min_{\fspcs{est}}\max\Bigg\{
\frac{\nlone{\fest}{f_k}}{\nlone{f_k}{\cP_{1,1}}},
\frac{\nlone{\fest}{f_k'}}{\nlone{f_k'}{\cP_{1,1}}}
\Bigg\},
\]
By the mirror image symmetry between $f_k$ and $f_k'$ about 1/2, $\fspcs{est}= u$ is the optimal estimate to within an additive $\cO(\epsilon/k)$.
This lower bounds $c$ as
\begin{align*}
\frac{c}{1-2\delta}\ge 
\frac{\nlone{f_k}{u}-\cO(\epsilon/k)}
{\nlone{f_k}{\cP_{1,1}}}&\overset{(a)}\ge \frac{\nlone{f_k}{u}-\cO(\epsilon/k)}
{\nlone{f_k}{g}}\\
&=\frac{\nloneint{[0,1/2]}{f_k}{u}+\nloneint{(1/2,1]}{f_k}{u}-\cO(\epsilon/k)}{\nloneint{[0,1/2]}{f_k}{g}+\nloneint{(1/2,1]}{f_k}{g}}\\
&\overset{(b)}{=}\frac{\|h_k\|_{1, [0,1/2]}+\nloneint{[0,1/2]}{g}{u}+\nloneint{(1/2,1]}{f_k}{u}-\cO(\epsilon/k)}
{\|h_k\|_{1, [0,1/2]}+\nloneint{(1/2,1]}{f_k}{g}}\\
&\overset{(c)}=\frac{\epsilon/4+\epsilon/4+\|f_k-u\|_{1, [1/2,1]}-\cO(\epsilon/k)}
{\epsilon/4+\nloneint{(1/2,1]}{f_k}{g}}\\
&\overset{(d)}=\frac{\epsilon/4+\epsilon/4+\|f_k-u\|_{1, [1/2,1]}-\cO(\epsilon/k)}
{\epsilon/4+\int_{1/2}^1 g(x)dx - \int_{1/2}^1 f_k(x)dx}\\
&=\frac{\epsilon/4+\epsilon/4+\|f_k-u\|_{1, [1/2,1]}-\cO(\epsilon/k)}
{\epsilon/4+1/2+\epsilon/4 - \int_{1/2}^1 f_k(x)dx}\\
&\overset{(e)}=\frac{\epsilon/4+\epsilon/4+\|f_k-u\|_{1, [1/2,1]}-\cO(\epsilon/k)}
{\epsilon/2+\cO(\epsilon/k)}\\
&\overset{(f)}{\ge}\frac{\epsilon/4+\epsilon/4+\epsilon/2-\cO(\epsilon/k)}
{\epsilon/2+\cO(\epsilon/k)}=2-\cO\Paren{\frac1k},
\end{align*}
where $(a)$ follows since $g\in \cP_{1,1}$, $(b)$ follows since $h_k$ is a discrete distribution, $(c)$ follows since $h_k$ has a total mass $\epsilon/4$ and since $\nloneint{[0,1/2]}{g}{u}=\epsilon/4$ by a straightforward calculation,
$(d)$ follows since $g\ge f_k$ in $(1/2,1]$, $(e)$ follows since the area of $f_k$ and $u$ on $I=(1/2,1]$ are equal to within an additive $\cO(\epsilon/k)$,
and $(f)$ follows since $\nloneint{(1/2,1]}{f_k}{u}
\ge 2\nloneint{(1/2,1]}{f_k}{g} 
-\cO(\epsilon/k)$.
Choosing $\delta\searrow 0$ and $k\nearrow \infty$ completes the proof.

% $f=f_0+f_2'$ as shown in Figure~\ref{ex2}. As $f_2'$ is a discrete sub-distribution and $f_0\in \cP_{1,1}$,
% $\nlone{f}{\cP_{1,1}}=\nlone{f}{f_0}=1/3$.
% As $n$ samples cannot distinguish between $f$ and 
% the symmetric \[v(x)=
% \begin{cases}
% 4/3-4/3 \cdot x & 0\le x<1/2\\
% 4/3\cdot x & 1/2\le x\le 1,
% \end{cases}\]
% % optimal approximation via the uniform $\fest = u$. This results in $c_{1,1}\ge \nlone{f}{u}/\nlone{f}{\cP_{1,1}}= (1/3+1/6)/(1/3)=3/2$.

% % Consider the functions $g(x) = 2/3\cdot (1+x)$
% % and
% % \[h(x)
% % \ed
% % \begin{cases}
% % 2/3\cdot (1+x) & 0\le x\le 1/2\\
% % 2/3\cdot (2-x) & 0\le x\le 1/2.
% % \end{cases}\]

% In other words $f_k(x)$ consists of $h(x)$ modified 
% for $x\in [1/2,1]$ to alternate between $0$ on  $J_i$ and $g(x)$ on $K_i$ for $i\in \sett{k/2}$. By construction, observe that the width $|J_i|\le |K_i|$. Therefore the best approximation to $f_k(x)$

% and consider $f \ed \lim_{k\rightarrow \infty } f_k$. It may be verified that $f$ is a distribution and that $\nlone{f}{\cP_{1,1}}$ is achieved by $g\in \cP_{1,1}$.
% % Repeat this construction by 
% % modifying $h(x)$ along $0\le x\le 1/2$ 
% As $n$ samples
% cannot be used to distinguish $f$ and $h$, by symmetry of $h$, 
% the uniform distribution $u\in \cP_{1,1}$ is an optimal estimator which implies 
% \[
% c_{1,1} \!\ge\! \frac{\nlone{f}{u}}{\nlone{f}{\cP_{1,1}}}
% \!=\!\frac{\nlone{f}{u}}{\nlone{f}{g}}\!=\!\frac{1/6+1/6+2/6}{1/6}\!=\!2.
% \]
\end{Proof}

% \blu{with positive spike best I can do is 3/2}
% Let $f_1(x) = 4/3\cdot x$ and let $f_2'$ be a discrete sub-distribution by assigning a probability of $1/(3n^3)$ on $n^3$ samples drawn according to $f_2(x)=4-8x, x\in [0,1/2]$. 
% Consider the distribution $f=f_1+f_2'$ as shown in Figure~\ref{ex2}. As $f_2'$ is a discrete sub-distribution and $f_1\in \cP_{1,1}$,
% $\nlone{f}{\cP_{1,1}}=\nlone{f}{f_1}=1/3$.
% As $n$ samples cannot distinguish between $f$ and 
% the symmetric \[v(x)=
% \begin{cases}
% 4/3-4/3 \cdot x & 0\le x<1/2\\
% 4/3\cdot x & 1/2\le x\le 1,
% \end{cases}\]
% optimal approximation via the uniform $\fest = u$. This results in $c_{1,1}\ge \nlone{f}{u}/\nlone{f}{\cP_{1,1}}= (1/3+1/6)/(1/3)=3/2$.

\subsection{Description and proof of Lemma~\ref{lem:minor}}
\label{pf:minor}
% The following lemma shows that any $c$-factor approximation for $\cP_{t,d}$ is also a $c$-factor approximation for any distribution class that is well approximated by $\cP_{t,d}$. 
The following lemma shows that if $\fest$ is a $c$-factor approximation for $\cP_{t,d}$ for some $t$ and $d$ and achieves the min-max rate of a distribution class $\cC$, then $\fest$ is also a $c$-factor approximation for $\cC$.
\begin{Lemma}
\label{lem:minor}
If $\fest$ is a $c$-factor approximation for $\cP_{t,d}$ and
for all $f$ in a class $\cC$, $c\cdot \nlone{f}{\cP_{t,d}}+\cO(\cR_n({\cP_{t,d})})\le \cO(\cR_n(\cC))$, then for any $f$, not necessarily in $\cC$,
\[
\nlone{\fest}{f} \le c\cdot \nlone{f}{\cC}+\cO(\cR_n(\cC)).
\]
% then $\fest$ is a $c$-factor approximation for $\cC$.
\end{Lemma}
\begin{Proof}
% Let $ \fest $ be a $ c $-factor approximation for $ \cP_{t,d} $. 
% Suppose $f$ is a distribution potentially outside of $\cC$.
For a distribution $g$ and class $\cD$, let
$g_{\cD}\in \cD$ be the closest approximation to $g$ from $\cD$, namely achieving $\nlone{g}{g_{\cD}}=\nlone{g}{\cD}$.
Then for any distribution $f$,
\begin{align*}
\lone{\fest-f} &\le c\cdot \nlone{f}{\cP_{t,d}} + \cO\Paren{ \cR_n(\cP_{t,d})}\\
&\overset{(a)}\le  c\cdot \lone{f-f_{\cC_{\cP_{t,d}}}
% q_{\cP_{t,d}}(q_{\cC}(f))
} +\cO\Paren{ \cR_n(\cP_{t,d})}\\
&\le c\cdot \lone{f-f_{\cC}}+ c\cdot \lone{f_{\cC_{\cP_{t,d}}}-f_{\cC}}+ \cO\Paren{ \cR_n(\cP_{t,d})}\\
&\overset{(b)}= c\cdot \nlone{f}{\cC}+ c\cdot \nlone{f_{\cC}}{\cP_{t,d}}+ \cO\Paren{ \cR_n(\cP_{t,d})}\\
&\overset{(c)}=c\cdot \nlone{f}{\cC} + \cO\Paren{\cR_n({\cC})},
\end{align*}
where in $ (a) $, just as $f_{\cC}$ is the $\cC$ distribution closest to $f$, $f_{\cC_{\cP_{t,d}}}$ is the $\cP_{t,d}$ distribution closest to $f_{\cC}$ and the inequality follows since $f_{\cC_{\cP_{t,d}}} \in \cP_{t,d} $ and by definition, $ \nlone{f}{\cP_{t,d}}$
is the least distance from $f$ to any $q\in \cP_{t,d}$,
$(b)$ follows since by definition, $f_{\cC_{\cP_{t,d}}}$ is the best approximation to $f_{\cC}$
from $\cP_{t,d}$, and $(c)$
follows from the property of $\cC$ considered in the lemma as $f_{\cC}\in \cC$.
\end{Proof}

\section{Proofs for Section~\ref{sec:singlep}}
\label{sec:appen3}

\subsection{Proof of Lemma~\ref{lem:poly}}
\label{ap:poly}
\begin{Proof}
%Here we describe a simplified proof that produces a larger 
%constant, namely $8\pi d + 2\sqrt{2}(d+1)$. But the proof can be improved (by considering a different $f_d$) to obtain $3d+\text{\scriptsize $\mathcal{O}$}(1)$.
%\\
Observe that for any $q\in \cP_d$ and interval $J$,
\[\Delta_J(q)
\ed \max_{x\in J} q(x) - \min_{x\in J} q(x)
\le \int_J |q'(t)|dt,
\]
where $q'$ denotes the first derivative of $q$.
The case of $d=0$ is trivial. We give a proof for $d\ge 1$. Consider
\[
f_d(x)
\, 
\ed
\,
\frac{1}{\text{\normalsize $x$}}
\sin(d \arcsin(x)).
\]
% {\bf Claim 1}\quad $f_d$ is a degree-$(d\!-\!1)$ polynomial for odd $d$. E.g.,
% $
% f_5(x)
% =
% 5 - 20 x^2 + 16 x^4.
% $
% \\
The following two claims (indicated with the overset (*)) may be verified via Wolfram Mathematica e.g. version 12.3.\\
{\bf Claim 1}\quad $\norm{f_d}_2\le \sqrt{\pi\,d}$,\,
because
\begin{align*}
\norm{f_d}_2^2
&\overset{(*)}=
2\int_0^1
\frac{\sin^2(d \arcsin(x))}{x^2}\,
\,dx\le
\int_{0}^{\frac{\pi}{2}}
\frac{1-\cos(2d\theta)}{ \sin^2 \theta}\, d\theta
\overset{}{=}\pi\, d.
\end{align*}\\
{\bf Claim 2}\quad
$f_d(0) \overset{(*)}= d$ and  $f'_d(0)\overset{(*)}=0$. \\
Let $f(t)\ed q(t)\sqrt{1-t^2}$ and note that for any $x\in [0,1]$,
\begin{align*}
\sqrt{1-x^2}\int_{-x}^x |q'(t)|\,dt
&\le
\int_{-1}^{1} |q'(t)|\sqrt{1-t^2}\,dt
\le
\int_{-1}^1 
|f'(t)|\,dt
+
\int_{-1}^1 
\frac{|t\, q(t)|}{\sqrt{1-t^2}},dt,
\end{align*}
where the final inequality follows by integration by parts.

For simplicity let $g_d(x) 
\ed
\sqrt{\frac{2}{\pi d}}
\cdot f_{\frac{d}{2}}(x)$, then  $g_d(0)=\sqrt{\frac{d}{2\pi}}$ and $g'_d(0)=0$. Hence,
\vspace{-.5em}
\begin{align*}
\frac{1}{2\pi}|f'(0)|
&=
\frac{1}{d}\,
|f'(0)|\,
g^2_d(0)
{\overset{(a)}{\le}}
2\max_{t} |f(t)|\,
g^2_d(t)
{\overset{(b)}{\le}}
4d\int_{-1}^{1} |q(t)|\, g^2_d(t)\, dt,
\end{align*}
where $(b)$ follows from Bernstein's inequality since  
$|q(t)|\sqrt{1-t^2}
\le d\int_{-1}^1 |q(z)|dz$ for a degree-$d$ polynomial $q$ and any $t\in [0,1]$, and applying the inequality to $q(t)g_d^2(t)$ of degree $2d-2$. For $(a)$ we apply Bernstein's inequality to 
$q(z)=\frac{d}{dz} \Paren{ f(z) g^2_d(z)}$, set $t=0$ and notice that $g_d'(0)=0$.
\\
Equivalently by a change of variables, for
$c\,\ed 8\pi$
and $f_{\text{\tiny T}}(\theta)\ed f(\sin \theta)$,
\begin{align*}
|f_{\text{\tiny T}}'(0)|=
|f'(0)|\le c \ d\int_{-1}^{1} |q(t)|\, g^2_d(t)\, dt
% &=c \, d\int_{-\pi}^{\pi} \Big|\frac{f(\sin \theta)}{|\cos \theta|}\Big|\,g^2_d(\sin\theta)\cos \theta d\theta
\le
c \, d\int_{-\pi}^{\pi} |f_{\text{\tiny T}}(\theta)|\,g^2_d(\sin\theta)d\theta.
\end{align*}
Using the same reasoning for 
$f_{\text{\tiny T}}(\theta + \alpha)
$ instead of $f_{\text{\tiny T}}(\theta )
$
we obtain
\[
|f_{\text{\tiny T}}'(\alpha)|
\le
c\, d\int_{-\pi}^{\pi} |f_{\text{\tiny T}}(\theta+\alpha)|\,g^2_d(\sin\theta)\,d\theta
=
c\, d\int_{-\pi}^{\pi} |f_{\text{\tiny T}}(\theta)|\,g^2_d(\sin(\theta-\alpha))\, d\theta,
\]
where the equality follows since 
$|f_{\text{\tiny T}}(\theta+\alpha)|\,g^2_d(\sin(\theta))$
is a periodic function with period $2\pi$.
Integrating both sides for $\alpha$ from $-\pi$ to $\pi$ yields
\[
\int_{-1}^1 
|f'(t)| \,dt
=
\int_{-\pi}^{\pi} |f_{\text{\tiny T}}'(\alpha)|\, d\alpha
\,{\overset{(a)}\le}\,
c\,d\int_{-\pi}^{\pi} |f_{\text{\tiny T}}(\theta)|
\,
d\theta
=
c\,d\int_{-1}^{1} |q(t)|
\,
dt,
\]
where (a) follows since $\norm{g_d}_2\le 1$ from Claim 1 and the definition of $g_d$.
Finally, 
for a proper $\tau\in(0, 1)$ of order $1/d^2$,
\begin{align*}
\int_{-1}^1 
\frac{|t\,q(t)|}{\sqrt{1-t^2}}\,dt
&\le 
\int_{-1+\tau}^{1-\tau}\frac{|q(t)|}{\sqrt{1-t^2}}\,dt
+
2\max_t |q(t)|\int_{1-\tau}^{1}\frac{t}{\sqrt{1-t^2}}\,dt\\
&{\overset{(a)}{\le} }
\frac{\int_{-1}^1 |q(t)|\,dt}{\sqrt{2\tau-\tau^2}}
+
2\max_t |q(t)|\int_{1-\tau}^{1}\frac{t}{\sqrt{1-t^2}}\,dt\\
&{= }
\frac{\int_{-1}^1 |q(t)|\,dt}{\sqrt{2\tau-\tau^2}}
+
2\max_t |q(t)|\cdot \sqrt{2\tau-\tau^2}\\
&{\overset{(b)}{\le} }
\frac{\int_{-1}^1 |q(t)|\,dt}{\sqrt{2\tau-\tau^2}}
+
2\sqrt{2\tau-\tau^2} 
\cdot (d+1)^2
\int_{-1}^1|q(t)|\, dt\\
&\overset{(c)}=
{2\sqrt{2}}
(d+1)\int_{-1}^1|q(t)|\, dt, 
\end{align*}
where $(a)$ follows since $1/\sqrt{1-t^2}\le 1/\sqrt{2\tau - \tau^2}$ for $t\in [-1+\tau,1-\tau]$, $(b)$ follows from the Markov Brothers' inequality, and $(c)$
holds for some $\tau = \cO(1/d^2)$.
Proof is complete from the fact that 
$8\pi + 2\sqrt{2}<28$.
\qedhere
\end{Proof}

\subsection{Proof of Lemma~\ref{lem:jajaja}}
\label{ap:jajaja}
Since for any $i\in \sett{m}$, $\barI_i^+ \in \barI^+$ and $\barI_i^- \in \barI^-$ both have $\lceil \ell (d+1)/2^{i/4}\rceil $ intervals
for $i\in \sett{m}$, it follows that the total number of
intervals in $(\barI^-,\barI^+)$ is upper bounded as
\begin{align*}
2\Paren{\sum_{i=1}^{m} |\barI_i| +1}
&= 2\Paren{\sum_{i=1}^{m}  \Bigg \lceil\frac{\ell(d+1)}{2^{i/4}}\Bigg \rceil +1}\\
&\le  2\Paren{\sum_{i=1}^{m}  \Paren{\frac{\ell(d+1)}{2^{i/4}}+1} +1}\\
&\le  2\Paren{\sum_{i=1}^{\infty}  \frac{\ell(d+1)}{2^{i/4}} +m+1}\\
&=2\Paren{\frac{\ell(d+1)}{2^{1/4}}
\cdot \frac{1}{1-2^{-1/4}}+\log_2(\ell(d+1)^2)+1}\\
&\overset{(a)}\le 2\Paren{\frac{\ell(d+1)}{2^{1/4}}
\cdot \frac{1}{1-2^{-1/4}}+2\log_2(\ell(d+1))+1}\\
&\overset{(b)}\le 2\Paren{\frac{\ell(d+1)}{2^{1/4}}
\cdot \frac{1}{1-2^{-1/4}}+\frac{2\ell(d+1)}{\log 2}}\\
&\overset{}=2\ell(d+1)\Paren{\frac{1}{2^{1/4}-1}+\frac{2}{\log 2}} \\
&\le \frac{4\ell(d+1)}{2^{1/4}-1}.
\end{align*}
where
$(a)$ follows since $\ell \ge 1$ and
$(b)$ follows from the identity that
for any $x\ge 1, \log(x)\le x-1$. 

\subsection{Proof of Lemma~\ref{lem:polyhist}}
\label{ap:polyhist}
\begin{Proof}
We provide a proof for $I=[-1,1]$ by considering $\barI = \parti{[-1,1]}{d}{k}$. An identical 
proof follows for any other interval $I$ as its partition $\parti{I}{d}{k}$ is obtained by a linear translation of $\parti{[-1,1]}{d}{k}$.
% From Equation~\eqref{eqn:324} we may use 
% % the relationship between 
% the absolute value of the integral of $p$ to
% upper bound the $\ell_1$ distance between $p$ and $\barp_{\barJ^{d,k}}$.
For $ i\in \{1\upto m\} $,
let \[ I_{i} \ed I_i^+ \cup I_i^-.\]
Similarly let ${E}_m \ed {E}_m^+\cup {E}_m^-$.
Applying Lemma~\ref{lem:poly} with $a={1-1/2^{i}}$, we obtain
\begin{align*}
% \Delta_{I_i}(p)\overset{(a)}
% \le
\int_{I_i}|p'(x)|dx \overset{(a)}\le 
\int_{-\Paren{1-1/2^{i}}}^{\Paren{1-1/2^{i}}} |p'
(x)|dx
&\overset{}{\le}
\frac{28(d+1)\loneint{I}{p}}{(1-\Paren{1-1/2^{i}}^2)^{1/2}}\\
&\le
\frac{28(d+1)\loneint{I}{p}}{(1-\Paren{1-1/2^{i}})^{1/2}}\\
&\le
28\cdot 2^{i/2}\cdot (d+1)\loneint{I}{p},
\end{align*}
where 
% $(a)$ follows as for the polynomial $p$,
% $\Delta_{I_i}(p)\overset{(a)} 
% =\max_{x\in I_i} p(x)
% -\min_{x\in I_i} p(x)
% \le 
% \int_{I_i}|p'(x)|dx
% $, 
$(a)$ follows since $I_i\subseteq [-\Paren{1-1/2^{i}}, \Paren{1-1/2^{i}}]$.
As $\barI_i^+$ consists of 
$\lceil\ell(d+1)/2^{i/4}\rceil$ equal width intervals from Equation~\eqref{eqn:l} and since $\barI_i^+$ is of width $|\barI_i^+| = 1/2^i$, it follows 
that each interval in $\barI_i^+$ (and similarly for $\barI^-$)
is of width $ \le 1/ (2^{3i/4}\cdot l(d+1))$.
Thus from Equation~\eqref{eqn:324},
the $\ell_1$ difference between $ p $ and $\barp_{\barI}$ over $I_i$ is given by
\begin{align*}
\nloneint{I_i}{p}{\barp_{\barI}}
&\le 
\sum_{J\in \barI_i}
\Delta_{J}(p)\cdot |J|\\
&\le 
\sum_{J\in \barI_i}
\Delta_{I_i}(p)\cdot \frac{1}{2^{3i/4}\cdot \ell(d+1)}\\
&\le 
\Paren{\sum_{J\in \barI_i}
\int_{J}|p'(x)|dx} \cdot \frac{1}{2^{3i/4}\cdot \ell(d+1)}\\
&\le 
\int_{I_i}|p'(x)|dx \cdot \frac{1}{2^{3i/4}\cdot \ell(d+1)}\\
&\le 
28\cdot 2^{i/2}\cdot (d+1)\loneint{I}{p}.\frac{1}{2^{3i/4}\cdot \ell(d+1)}
=
\frac{28\cdot 2^{-i/4}\loneint{I}{p}}{\ell}.
\end{align*}
% Since $p$ and $\bar{p}_{\barJ}$ have the same mass in $I$, the their $\tvv$ distance within $I$ can be related to the $\ell_1$ distance as
Therefore
\begin{align*}
\nloneint{I}{p}{\barp_{\barI}}
&=\sum_{i=1}^{m} \nloneint{I_i}{p}{\barp_{\barI}} + \nloneint{{E}_{m}}{p}{\barp_{\barI}}\\
&\le 
\sum_{i=1}^{m}
\frac{28\cdot 2^{-i/4}\loneint{I}{p}}{\ell}
+\max_{x\in {E}_m}{p(x)}\cdot \frac{2}{\ell(d+1)^2}\\
&\overset{(a)}{\le }
\sum_{i=1}^{m}
\frac{28\cdot 2^{-i/4}\loneint{I}{p}}{\ell}
+(d+1)^2\loneint{I}{p}\frac{2}{\ell(d+1)^2}\\
&\overset{(b)}{\le }
\frac{\loneint{I}{p}}{\ell}
\Paren{\frac{28}{1-2^{-1/4}}+2}\\
&\overset{(c)}{\le }
\frac{4(d+1)\loneint{I}{p}}{k(2^{1/4}-1)}
\Paren{\frac{28}{1-2^{-1/4}}+2}\\
% &\overset{(c)}=\frac{28\cdot 2^{1/4}\loneint{I}{p}(d+1)}{k(2^{1/4}-1)^2}
&\overset{}{\le } 
\frac{3764(d+1)\loneint{I}{p}}{k},
\end{align*}
where $(a)$ follows since $I$ is a symmetric interval, $\forall x\in I$, $p(x)\le (d+1)^2\loneint{I}{p}$ from the Markov Brothers' inequality,
$(b)$ follows from the infinite negative geometric sum and $ (c) $
follows since $\ell\ed  k(2^{1/4}-1)/(4(d+1)) $ as defined in Equation~\eqref{eqn:l}.
\end{Proof}

\subsection{Proof of Lemma~\ref{lem:diff2}}
\label{ap:diff2}
\begin{Proof}
Select a $p^*\in \cP_d$ that achieves 
$\nloneint{I}{f}{p^*}=\nloneint{I}{f}{\cP_d}$. Then
\begin{align*}
\nloneint{I}{\fspl}{f}&\le \nloneint{I}{\fspl}{p^*}+\nloneint{I}{p^*}{f}\\
&\overset{(a)}\le 
2\cdot \nloneint{I}{f}{p^*}+\frac{c_1(d+1)\nloneint{I}{p^*}{\fpoly}}k +\akkint{k}{I}{\femp-f}
\\
&\le 
2\cdot \nloneint{I}{f}{p^*}+\frac{c_1(d+1)\Paren{\nloneint{I}{p^*}{f}+\nloneint{I}{f}{\fpoly}}}k +\akkint{k}{I}{\femp-f}\\
&\overset{(b)}{\le} 
\Paren{2+\frac{c_1(d+1)(1+c')}{k}}\nloneint{I}{f}{p^*}
+\Paren{\frac{c_1(d+1)c''}{k}+1}
\akkint{k}{I}{\femp -f}
+\frac{c_1(d+1)}{k}\cdot{\eta},
\end{align*}  
where $(a)$ follows from 
setting $p=p^*$ in
Lemma~\ref{lem:diff}, and
$(b)$ follows from using 
$\nloneint{I}{\fpoly}{f}\le c' \nloneint{I}{f}{p^*}+c''\akkint{d+1}{I}{\femp-f}+\eta$ along with the fact that $k\ge d+1$ (so that $\akkint{d+1}{I}{\femp-f}\le 
\akkint{k}{I}{\femp-f}$).
\end{Proof}

\subsection{Proof of Theorem~\ref{thm:singlep}}
\label{ap:singlep}
\begin{Proof}
Since $c=3$ and $c'=2$ for the $\fadls$ estimate,
it follows for $I=[X_{(0)},X_{(n)}]$
from Lemma~\ref{lem:diff2} that
\begin{align*}
\nloneint{I}{\fspl}{f}
&\le 
\Paren{2+\frac{3c_1(d+1)}{k}}\nlone{f}{p^*}
+\Paren{\frac{2c_1(d+1)}{k}+1}
\akk{d+1}{\femp -f}
+\frac{c_1(d+1)}{k}\cdot{\eta_d}\\
&\le 
\Paren{2+\gamma}\nlone{f}{\cP_d}
+\Paren{\frac{\gamma}{4}+1}
\akk{d+1}{\femp -f}
+\frac{\gamma}{8}\cdot{\eta_d}
\end{align*}
where the last inequality follows since we choose $k = k(\gamma)  \ge  8c_1(d+1)/\gamma $. Let $J=\reals\setminus I$.
Using Lemma~\ref{lem:vc},  
\begin{align*}
\E \nlone{\fspl}{f}
&\le 
\Paren{2+\gamma}\nloneint{I}{f}{\cP_d}
+\Paren{\frac{\gamma}{4}+1}
\E \akk{d+1}{\femp -f}
+\frac{\gamma}{8}\cdot{\eta_d}
+\Paren{2+\gamma}\loneint{J}{f}\\
&=
\Paren{2+\gamma}\nloneint{I}{f}{\cP_d}
+\Paren{\frac{\gamma}{4}+1}
\E \akk{d+1}{\femp -f}
+\frac{\gamma}{8}\cdot{\eta_d}
+\Paren{2+\gamma}\E \akkint{2}{J}{f-\femp}\\
&\overset{(a)}\le 
\Paren{2+\gamma}\nloneint{I}{f}{\cP_d}
+\Paren{\frac{\gamma}{4}+1}
\cO\Paren{\sqrt{\frac{k}{n}}}
+\frac{\gamma}{8}\cdot{\eta_d}
+3\cdot \cO\Paren{\sqrt{\frac{2}{n}}}\\
&\overset{(b)}\le 
\Paren{2+\gamma}\nlone{f}{\cP_d}
+
\cO\Paren{\sqrt{\frac{d+1}{\gamma\cdot n}}},
\end{align*}
where $(a)$ follows since $\gamma<1$ and from Lemma~\ref{lem:vc},
and $(b)$ follows as $\eta_d = \sqrt{(d+1)/n}$, $k=\cO(d+1)$ and $0<\gamma<1$.
\end{Proof}

\section{Proofs for Section~\ref{sec:multip}}
\label{sec:appen4}
\subsection{Proof of Theorem~\ref{thm:main}}
\label{ap:main}
From Lemma~\ref{lem:almostmain}, 
\begin{align*}
\nlone{\fout_{t,d,\alpha}}{f} &\le
\Paren{2+\frac{4c_1(d+1)}{k}
+\frac{1+k/(d+1)}{\beta-1}}\nlone{f}{\cP_{t,d}}+\Paren{3+2c_1+\frac{2}{\beta-1}}\akk{2\beta t\cdot k }{\femp-f}\\
&\phantom{\le}+\Paren{\frac{c_1(d+1)}{k}+\frac{k}{(\beta-1)(d+1)}}\eta_d\\
&\overset{(a)}\le
\Paren{2+\frac{4c_1(d+1)}{k}
+\frac{\alpha(d+1)}{4k}\cdot\Paren{1+\frac{k}{d+1}}
}\nlone{f}{\cP_{t,d}}+\Paren{3+2c_1+\frac{\alpha(d+1)}{2k}}\akk{2\beta t\cdot k }{\femp-f}\\
&\phantom{\le}+\Paren{\frac{c_1(d+1)}{k}+
\frac{\alpha(d+1)}{4k}\cdot \frac{k}{d+1}}\eta_d\\
&\overset{(b)}\le 
\Paren{2+\frac\alpha2+\frac\alpha2}\nlone{f}{\cP_{t,d}}+
\Paren{3+2c_1+\frac{\alpha}{2}}\akk{2\beta t\cdot k }{\femp-f}+
\Paren{\frac{\alpha}{8}+\frac{\alpha}{4}}\eta_d
\end{align*}
where $(a)$ follows since by definition
$\beta -1 = {4k}/(\alpha(d+1))$, $(b)$ follows since
$k\ed \lceil {8c_1(d+1)}/{\alpha} \rceil$ and since $0<\alpha<1$, $c_1>1$ imply
$k\ge d+1$. From Lemma~\ref{lem:vc},
\begin{align*}
\E \nlone{\fout_{t,d,\alpha}}{f}
&\le 
\Paren{2+\alpha}\nlone{f}{\cP_{t,d}}+
\Paren{3+2c_1+\frac\alpha2}\E\akk{2\beta t\cdot k }{\femp-f}+
\frac{3\alpha\eta_d}{8}\\
&\overset{}\le \Paren{2+\alpha}\nlone{f}{\cP_{t,d}}+
\Paren{3+2c_1+\frac\alpha2}\cO\Paren{\sqrt{\frac{2\beta t k}{n}}}+
\frac{3\alpha\eta_d}{8}\\
&\overset{(a)}\le \Paren{2+\alpha}\nlone{f}{\cP_{t,d}}+
\Paren{3+2c_1+\frac\alpha2}\cO\Paren{\sqrt{\frac{k^2t}{\alpha(d+1)}}}+
\frac{3\alpha\eta_d}{8}\\
&\overset{(b)}\le \Paren{2+\alpha}\nlone{f}{\cP_{t,d}}+
\Paren{3+2c_1+\frac\alpha2}\cO\Paren{\sqrt{\frac{c_1^2(d+1)^2t}{\alpha^3(d+1)n}}}+
\frac{3\alpha\eta_d}{8}\\
&\overset{(c)}\le \Paren{2+\alpha}\nlone{f}{\cP_{t,d}}+
\cO\Paren{\sqrt{\frac{t(d+1)}{\alpha^3 n}}},
\end{align*}
where $(a)$, $(b)$ both follow from the definitions of $\beta$, $k$ in 
Equations~\eqref{eqn:beta},~\eqref{eqn:k} 
and $(c)$ follows since $n_d\ed \sqrt{(d+1)/n}$ and $0<\alpha <1$.

\subsection{Proof of Lemma~\ref{lem:almostmain}}
\label{sec:almostmain}
For simplicity denote $\fout \ed \fout_{t,d,\alpha}$ and consider a 
particular $p^*\in \cP_{t,d}$ that achieves $\nlone{f}{\cP_{t,d}}$.
Let $\bar{F}$ denote the set of intervals in $\bar{I}_{\ADLS}$ that has $p^*$ as a single piece polynomial in $I$.
Let $\barJ\ed \bar{I}_{\ADLS} \setminus\bar{F}$ be the remaining intervals where $p^*$ has more than one polynomial piece. Since $p^*\in \cP_{t,d}$ has $t$ polynomial pieces, the number of intervals in $\barJ$ is $\le t$.

Recall that for any subset $S\subseteq \reals$, and integrable functions $g_1$, $g_2$, and integer $m\ge 1$,
% \newbla{okay to use $k$, or some other letter?}
$\nloneint{S}{g_1}{g_2}$, $\akkint{m}{S}{g_1-g_2}$ denote the $\ell_1$ and $\cA_m$  distances over $S$ respectively.
Equation (14) in~\cite{jay17} shows that over $\bar{F}$,
\begin{align}
\label{eqn:576}
\sum_{I\in \bar{F}} \nlone{\fadls_I}{f} \le& \ 3\nloneint{\bar{F}}{f}{p^*} + 2||\femp-f||_{\cA_{|\bar{F}|\cdot(d+1)},\bar{F}} +\eta_d.
\end{align}
We bound the error in $ \bar{F}$
by setting $p=p^*, \ \fpoly = \fadls_I$ 
in Lemma~\ref{lem:diff}, using Equation~\eqref{eqn:576}, and noting that $k\ge d+1$ as $c_1\ge 1$ from Equation~\eqref{eqn:k}:
\ignore{Self reference: Basically involves 2 steps of TEQ 
after selecting $p=p^*$ in that Lemma (and I think $\ADLS$
estimate is used as $\fpoly$ in the Lemma)
}
\begin{align}
\sum_{I\in \bar{F}} \nloneint{I}{\fout}{f}&\le 2\sum_{I\in \bar{F}} \nloneint{I}{f}{p^*}+||\femp -f||_{\cA_{k\cdot|\bar{F}|}}
+\frac{c_1(d+1)}{k}(\nlone{p^*}{f}+\nlone{f}{\fadls_I})\nonumber\\
&\le  \  2 \nloneint{\bar{F}}{f}{p^*}+(1+2c_1)||\femp -f||_{\cA_{k\cdot|\bar{F}|}}+\frac{c_1(d+1)}{k}\Paren{4\nlone{f}{p^*}
+\eta_d}.\label{eqn:cd}
\end{align}
From Lemma 49~\cite{jay17}, for all intervals $I\in \barJ$, the following Equation~\eqref{eqn:bccc} holds that they use to derive Equation~\eqref{eqn:bc}.
\begin{align}
\akkint{d+1}{I}{\fadls_I-\femp}
\label{eqn:bccc}
&\le \frac{\nlone{f}{p^*}+\akk{{2\beta t\cdot (d+1)}}{\femp-f}+\eta_d}{(\beta-1)t}.
\end{align}
\begin{align}
\sum_{I\in \barJ} \nloneint{I}{\fadls_I}{f} &\le \frac{\nlone{f}{p^*}+\akk{2\beta t\cdot(d+1)}{\femp-f}}{\beta-1} + 2\akk{2\beta t\cdot(d+1)}{\femp-f}+{2}\nloneint{\barJ}{f}{p^*}+\frac{\eta_d}{2(\beta-1)}.\label{eqn:bc}
\end{align}
Recall that we obtain $\fout$
% in interval $I$ via the $\SPLIT$ 
% routine that 
by adding a constant to $\fadls_I$ along each interval $I\in \parti{I}{d}{k}$
to match its area to $\femp$ in that interval.
Since 
$\parti{I}{d}{k}$ has $\le k$ intervals (Lemma~\ref{lem:jajaja}),
$\forall I \in \barJ$, 
\begin{align}
\nonumber
\nloneint{I}{\fout}{\fadls_I}
&\le \akkint{k}{I}{\fadls_I-\femp}\\
\label{eqn:bddd}
&\le \frac{k}{d+1}\akkint{d+1}{I}{\fadls_I-\femp}
\end{align}
where the last inequality follows from Property~\ref{prop:2}.

\ignore{
Upon applying Lemma~\ref{lem:diff} 
with $p = \fpoly=\fadls_I$ over $I\in \barJ$,
\[
\nloneint{I}{\fout}{\fadls_I}\le 
\nloneint{I}{f}{\fadls_I} +||\femp-f||_{\cA_k, I}.
\]
% \tcr{What happened to middle term in above? Are we applying $\fout$
% on $\ADLS$ output? Yep, seems like it. In that case it vanishes since $\fpoly=p_{\ADLS}^J$.}
Summing over all intervals in $\barJ$ and using $k\ge d+1$,
\begin{align}
\sum_{I\in \barJ}& \nloneint{I}{\fout}{\fadls_I} \le \sum_{I\in \barJ}\nloneint{I}{f}{\fadls_I}+\akk{k|\bar J|} {\femp-f}\nonumber\\
\le &  \ \frac{\nlone{f}{p^*}+\akk{2\beta t\cdot(d+1)}{\femp-f}}{\beta-1} + 3\akk{2\beta t\cdot k}{\femp-f}\nonumber\\
&+\nloneint{\barJ}{f}{p^*}
+\frac{\eta_d}{2(\beta-1)}.\label{eqn:ab}
\end{align}
}
Adding  
Equations~\eqref{eqn:bc}~and~\eqref{eqn:bddd} over intervals in $\barJ$
by noting $\barJ$ has $\le t$ intervals and $k\ge d+1$ implies
\begin{align}
\nonumber
\sum_{I\in \barJ} \nloneint{I}{\fout}{f}\le&
{\frac{1+k/(d+1)}{\beta-1}}\cdot \nlone{f}{p^*}
+2\nloneint{\barJ}{f}{p^*}
+2\Paren{1+\frac{1}{\beta-1}}
\akk{2\beta t\cdot k}{\femp-f}
\\
&+\frac{k\cdot \eta_d}{(d+1)(\beta-1)}
\label{eqn:barJ}.
\end{align}
Adding Equations~\eqref{eqn:cd} and~\eqref{eqn:barJ} proves Lemma~\ref{lem:almostmain}
(since $p^*$ satisfies $\nlone{f}{p^*}=\nlone{f}{\cP_{t,d}}$).

\section{Proofs for Section~\ref{sec:cv}}
\label{sec:appen5}
\subsection{Proof of Theorem~\ref{thm:middle}}
\label{ap:middle}
\begin{Proof}
Applying the probabilistic version of the VC inequality, i.e. Lemma~\ref{lem:vc}, (see \cite{dev12}) to Lemma~\ref{lem:almostmain} we have with probability $\ge 1-\delta$,
\[
\nlone{\fout_{t,d,\alpha}}{f}
\le 
c\cdot \nlone{f}{\cP_{t,d}}+
\cO\Paren{
\sqrt{\frac{t(d+1)+\log 1/\delta}{n}}}. 
\]
From the union bound, the above condition holds true for the $\log n$ sized estimate collection
$\{\fout_{t,d,\alpha}: t\in \{1,2,4\upto n\}\}$ with probability $\ge 1-\delta\cdot \log n$. 
Apply the method discussed in Section~\ref{sec:cvconstruct} with $\gamma = 2+2/\beta $, to obtain
$\tspcs{est} = \tspcs{est}_{\beta}$.
Using Lemma~\ref{lem:cvini} that
w.p. $\ge 1-\delta\cdot \log n$,
\begin{align*}
\lone{\fout_{\tspcs{est},d,\alpha}-f} &\overset{}\le \min_{t\in \{1,2,4\upto n\},d} \Paren{\Paren{1+\frac2{\gamma -2}}\cdot  c\cdot \nlone{f}{\cP_{t,d}} + (\gamma+ 1)\chi  {\sqrt{\frac{t(d+1)}{n}}}}\\
&\overset{(a)}\le \min_{0\le t\le n,d} \Paren{\Paren{1+\frac2{\gamma -2}}\cdot  c\cdot \nlone{f}{\cP_{t,d}} + \sqrt{2}(\gamma+ 1)\chi  {\sqrt{\frac{t(d+1)+\log 1/\delta}{n}}}}\\
&\overset{(b)}\le \min_{0\le t\le n,d} \Paren{\Paren{1+\beta}\cdot  c\cdot \nlone{f}{\cP_{t,d}} +   \cO\Paren{{\sqrt{\frac{t(d+1)+\log 1/\delta}{\beta^2 n}}}}},
\end{align*}
where $(a)$ follows from the fact that for any 
$1\le t\le n$, $\exists t'\in \{1,2,4\upto n\}: t'\in [t,2t]$ (so that $\nlone{f}{\cP_{t',d}}\le \nlone{f}{\cP_{t,d}}$ and $(b)$ follows since
$\gamma = 2+2/\beta $.
\end{Proof}

\subsection{Proof of Lemma~\ref{lem:cvini}}
\label{ap:cvini}
\begin{Proof}
For $i\ge i_\gamma$, from the triangle inequality, and as by definition, 
$\dist{v_{i_\gamma}}{v_i}\le \gamma c_i$ for all $i\ge i_\gamma$, 
\begin{align*}
\label{eqn:6}
\dist{v_{i_\gamma}}{v}&\le \dist{v_{i}}{v}+ \dist{v_{i_\gamma}}{v_{i}}
\le b_{i}+c_i + \gamma c_i 
= b_{i}+(1+\gamma) c_i.
\end{align*}

For $i< i_\gamma$, if 
\[
b_{i_\gamma - 1} \ge \frac{\gamma-2}{2}c_{i_\gamma},
\]
the proof follows since for any $1\le j'\le i_\gamma-1$,
\begin{align*}
\dist{v_{i_\gamma}}{v}\le b_{i_\gamma}+c_{i_\gamma}
\le 
b_{i_\gamma}+\frac{2}{\gamma-2}b_j
\overset{(a)}\le 
\Paren{1+\frac{2}{\gamma-2}}b_{j'},
\end{align*}
where $(a)$ follows since
$j'\le j<i_{\gamma}$.
On the other hand if 
\[
b_{i_\gamma - 1} < \frac{\gamma-2}{2} c_{i_\gamma},
\]
then $\forall j''\ge j + 1 $,
\[
\dist{v_j}{v_{j''}}
\le b_j +b_{j''}+ c_j 
+c_{j''}
\overset{(a)}\le 2b_j+ 2c_{j''} 
\le 2\cdot \frac{\gamma-2}{2}c_{i_\gamma} + 2c_{j''}\overset{(b)}\le \gamma c_{j''},
\]
where $(a)$ follows since
$j''\ge j$, and $(b)$ follows since $j''\ge j= i_\gamma$, contradicting the definition of $i_{\gamma}$.

\end{Proof}

\section{Additional Experiments}
\begin{figure*}[!ht]
\centering
\subfigure{\includegraphics[scale=1.1]{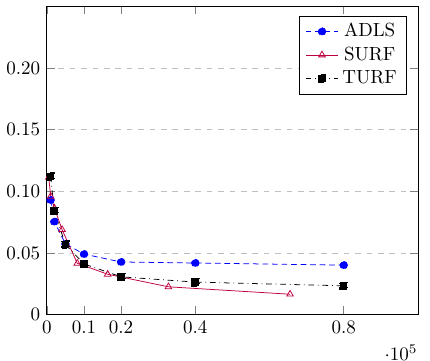}\label{aaaa}}\qquad
\subfigure{\includegraphics[scale=1.1]{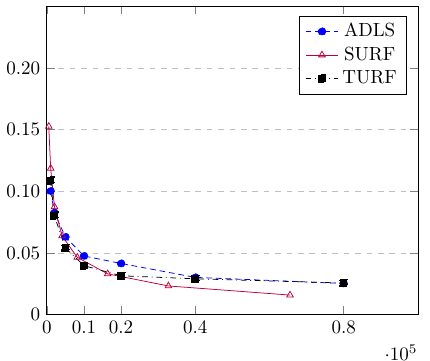}\label{bbbb}}\qquad
\subfigure{\includegraphics[scale=1.1]{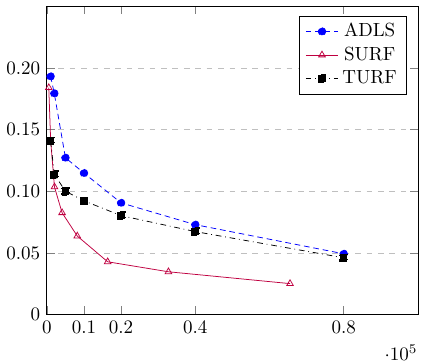}\label{cccc}}
\vspace{-1em}
\caption{$\ell_{1} $ error versus number of samples on the Beta, Gamma, and Gaussian mixtures respectively in Figure~\ref{plot:compare0}
for $d=1$.}
\label{plot:surf}
\end{figure*}
We compare $\SURF$~\cite{hao2021surf}
against $\TURF$ and $\ADLS$~\cite{jay17} for the non-noisy distributions considered in Section~\ref{sec:experi}, namely mixtures of Beta: $.4\text{B}(.8, 4)+.6\text{B}(2, 2) $,
Gamma: $.7\Gamma(2, 2)+.3\Gamma(7.5, 1) $, and Gaussians: .65$\cN$(-.45,$.15^2$)+.35$\cN$(.3,$.2^2$). This is shown in Figure~\ref{plot:surf}. While $\SURF$ achieves a lower error, this may be due to its implicit cross-validation method, unlike in $\ADLS$ and $\TURF$ that relies on our independent cross-validation procedure in Section~\ref{sec:cv}. While the primary focus of our work
was in determining the optimal approximation constant,
evaluating the experimental performance of the various piecewise polynomial estimators may be an interesting topic for future research.
\label{sec:appen6}

\end{document}